\documentclass[lettersize,journal]{IEEEtran}
\usepackage{amsmath,amsfonts}
\usepackage{algorithmic}
\usepackage{algorithm}
\usepackage{array}
\usepackage[caption=false,font=normalsize,labelfont=sf,textfont=sf]{subfig}
\usepackage{textcomp}
\usepackage{stfloats}
\usepackage{url}
\usepackage{verbatim}
\usepackage{graphicx}
\usepackage{cite}
\usepackage{threeparttable}
\usepackage{cuted}
\usepackage{subcaption}
\usepackage{tikz}
\usepackage{multirow}
\usepackage{xcolor}
\usepackage{booktabs}
\usepackage{svg}
\usepackage{tabularx}
\usepackage[normalem]{ulem}
\useunder{\uline}{\ul}{}
\begin{document}

\title{GLaD: Geometric Latent Distillation for \\Vision-Language-Action Models}
\author{%
Minghao Guo, Meng Cao, Jiachen Tao, Rongtao Xu, Yan Yan, Xiaodan Liang, Ivan Laptev, Xiaojun Chang
\thanks{Minghao Guo, Rongtao Xu, Xiaodan Liang, Ivan Laptev and Xiaojun Chang are with MBZUAI, Abu Dhabi, United Arabic Emirates (email: {minghao.guo, rongtao.xu, xiaodan.liang, ivan.laptev, xiaojun.chang}@mbzuai.ac.ae).}
\thanks{Meng Cao is with MBZUAI, Abu Dhabi, United Arabic Emirates (email: mengcaopku@gmail.com).}
\thanks{Jiachen Tao and Yan Yan are with University of Illinois Chicago, Chicage, United States (email: {jtao26, yyan55}@uic.edu).}
}



\maketitle

\begin{abstract}

Most existing Vision-Language-Action (VLA) models rely primarily on RGB information, while ignoring geometric cues crucial for spatial reasoning and manipulation.
In this work, we introduce GLaD, a geometry-aware VLA framework that incorporates 3D geometric priors during pretraining through knowledge distillation.
Rather than distilling geometric features solely into the vision encoder, we align the LLM's hidden states corresponding to visual tokens with features from a frozen geometry-aware vision transformer (VGGT), ensuring that geometric understanding is deeply integrated into the multimodal representations that drive action prediction.
Pretrained on the Bridge dataset with this geometry distillation mechanism, GLaD achieves 94.1\% average success rate across four LIBERO task suites, outperforming UniVLA (92.5\%) which uses identical pretraining data.
These results validate that geometry-aware pretraining enhances spatial reasoning and policy generalization without requiring explicit depth sensors or 3D annotations.


\end{abstract}

\section{Introduction}

\IEEEPARstart{V}{ision-language-action (VLA)} models have emerged as a promising paradigm for embodied intelligence, enabling robots to generate control actions directly from visual observations and natural language instructions. Recent works~\cite{kim2024openvlaopensourcevisionlanguageactionmodel, black2024pi0visionlanguageactionflowmodel, intelligence2025pi05visionlanguageactionmodelopenworld, intelligence2025pi06vlalearnsexperience} have demonstrated impressive performance on diverse manipulation tasks by leveraging large-scale multimodal pretraining. These models typically combine powerful vision encoders~\cite{radford2021learningtransferablevisualmodels, oquab2024dinov2learningrobustvisual, zhai2023sigmoidlosslanguageimage} and large language models to learn generalizable visuomotor policies from extensive robot demonstration datasets. 

\begin{figure}[t]
\centering
\begin{minipage}{0.9\linewidth}   
  \centering

  \begin{minipage}[c]{0.48\linewidth}
    \centering
    \includegraphics[width=\linewidth]{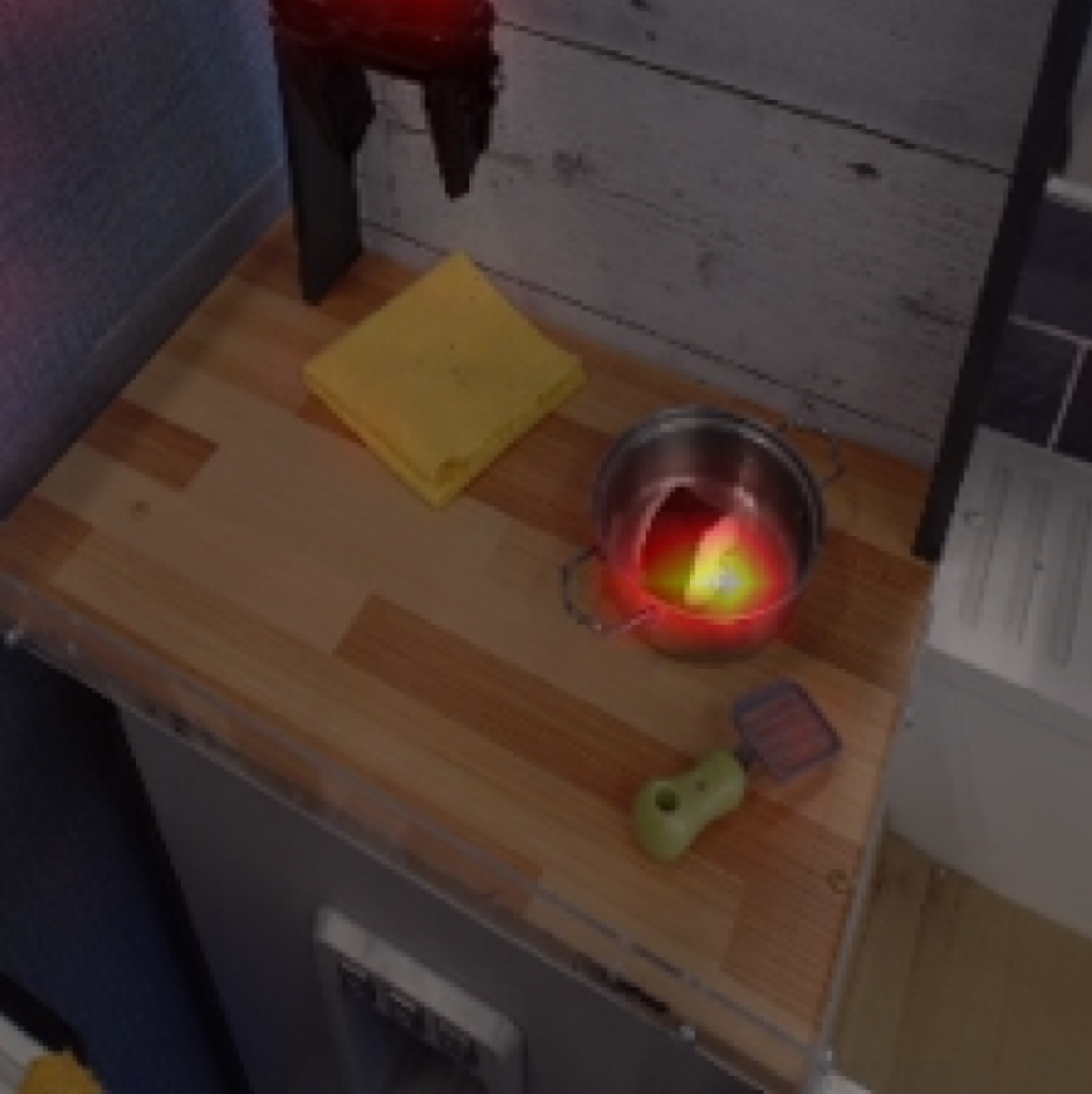}
  \end{minipage}
  \hfill
  \begin{minipage}[c]{0.48\linewidth}
    \centering
    \includegraphics[width=\linewidth]{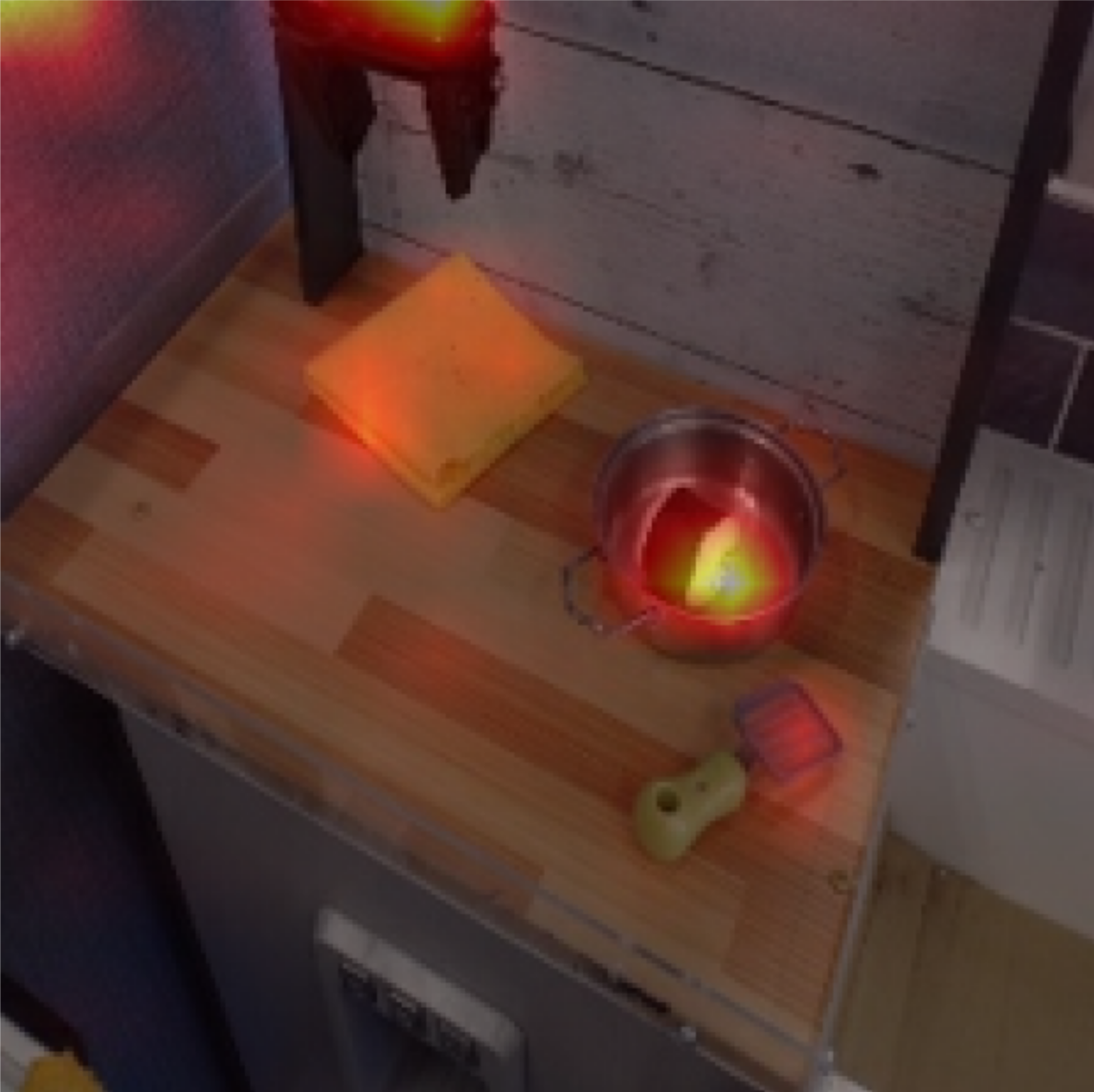}
  \end{minipage}

  \vspace{0.5em} 

  \begin{minipage}[c]{0.48\linewidth}
    \centering
    \includegraphics[width=\linewidth]{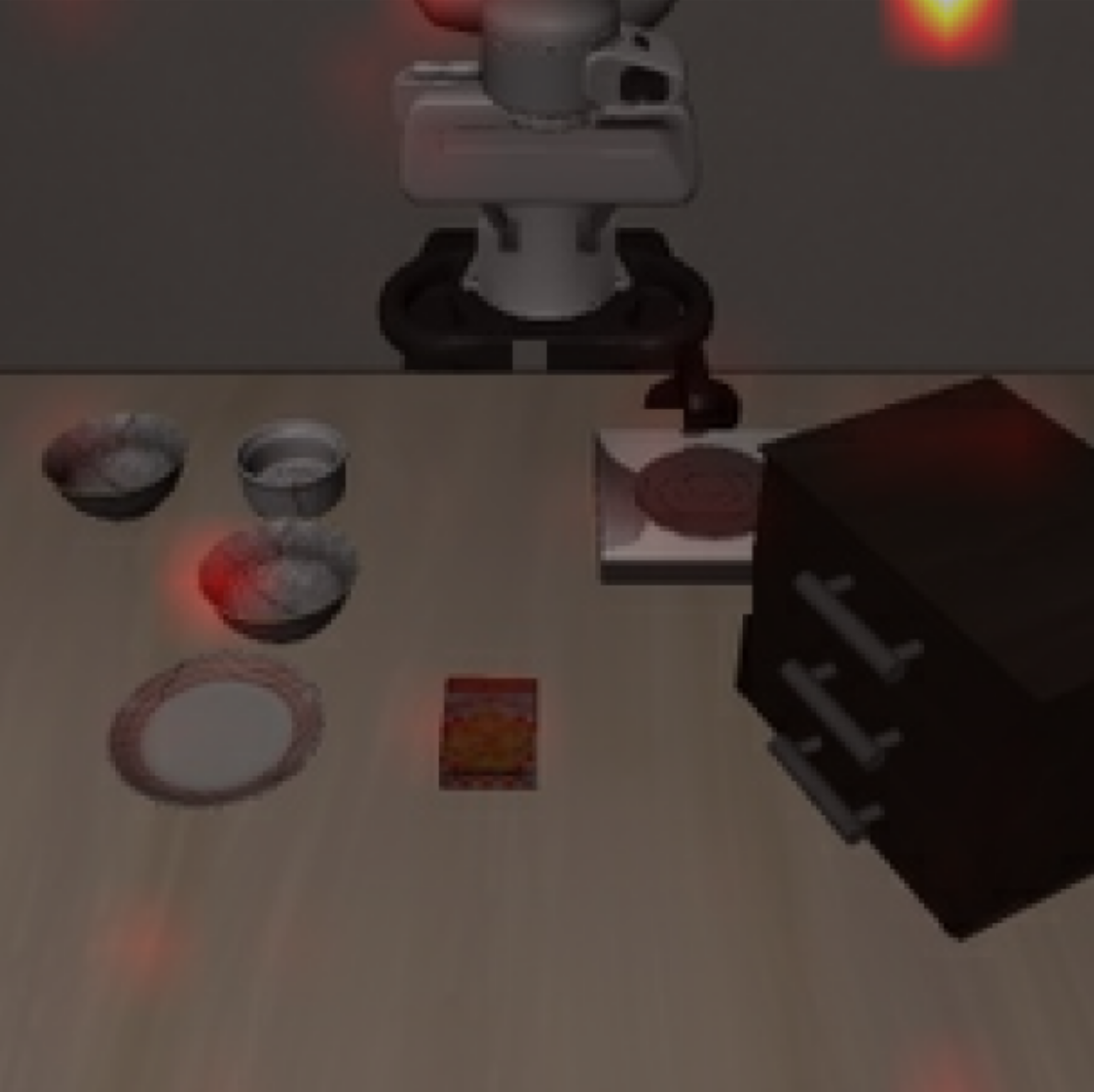}
    \caption*{VLA w/o geometry}
  \end{minipage}
  \hfill
  \begin{minipage}[c]{0.48\linewidth}
    \centering
    \includegraphics[width=\linewidth]{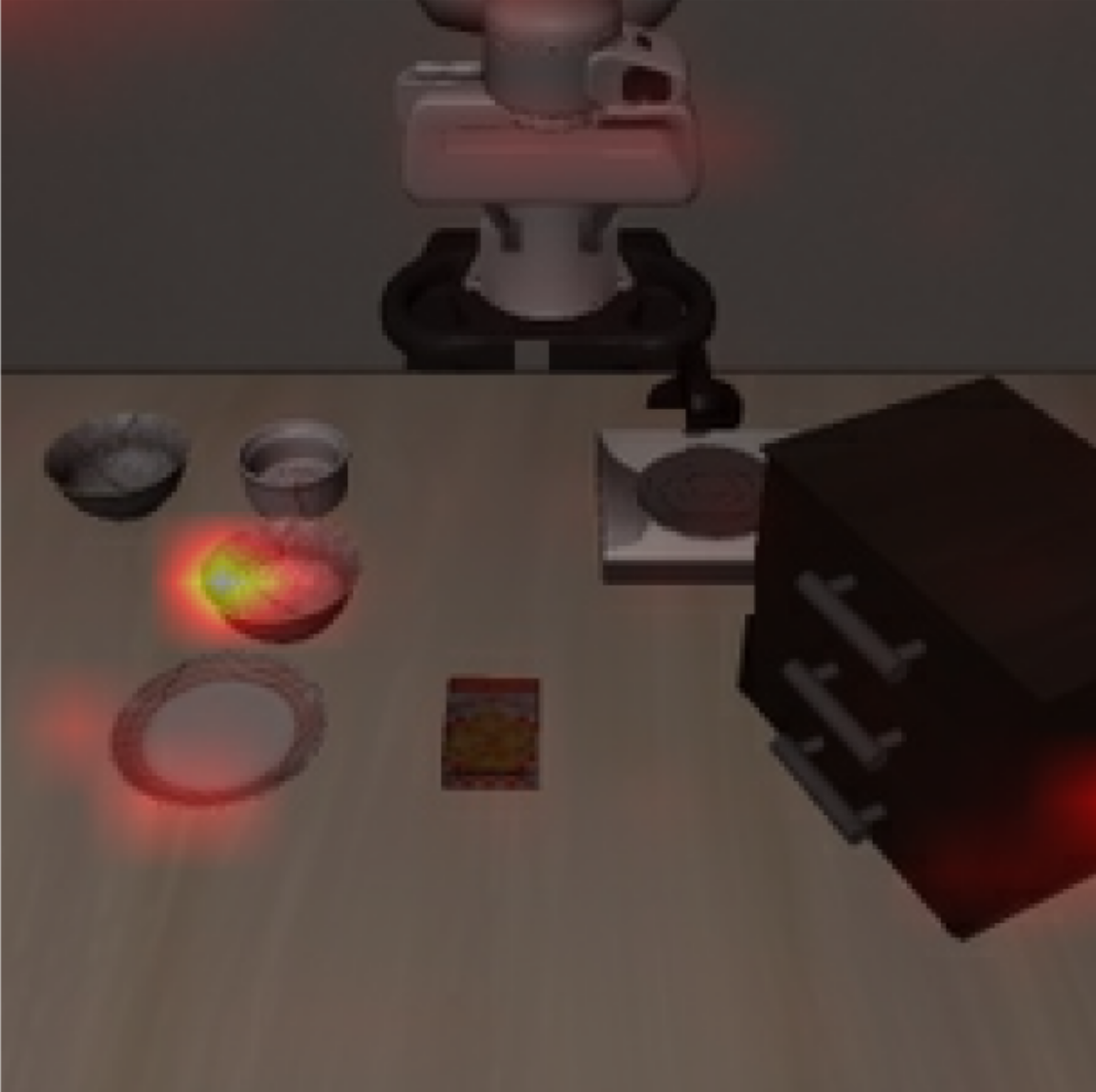}
    \caption*{VLA w/ geometry}
  \end{minipage}

\end{minipage}

\caption{Attention maps of VLA. Up: (Bridge scene) Move the table cloth from corner to edge of the table. Down: (LIBERO scene) Pick up the black bowl between the plate and the ramekin and place it on the plate.}
\label{fig:intro-attn}
\end{figure}

Despite these advances, current VLA architectures fundamentally lack \textit{geometric understanding}, which represent the capability of perceiving spatial positions, 3D structures, and relational arrangements among objects in a scene—knowledge that is essential for robots to reason about where objects are, how they relate to each other, and how to interact with them effectively. Most VLAs rely on vision encoders pretrained with 2D contrastive objectives such as CLIP~\cite{radford2021learningtransferablevisualmodels} or SigLIP~\cite{zhai2023sigmoidlosslanguageimage}, which excel at capturing semantic correspondences between images and text but do not encode 3D spatial information. These 2D embeddings represent visual scenes as flat semantic patterns without explicitly modeling depth, object poses, or spatial relationships—information that is critical for manipulation tasks where precise positioning matters, thus resulting wrong attention of the objects in the scene shown in Fig.~\ref{fig:intro-attn}. \IEEEpubidadjcol This raises a critical question: \textit{Can we inject geometric priors into VLA pretraining to enhance scene understanding and improve policy generalization?}

To address this challenge, we propose \textbf{GLaD}, \textbf{G}eometric \textbf{La}tent \textbf{D}istillation vision-language-action framework that incorporates 3D geometric knowledge. Our key insight is that integrating geometric priors through knowledge distillation can substantially enhance a VLA's ability to understand scene structure and reason about manipulation tasks. Specifically, GLaD introduces a geometry distillation mechanism during pretraining: we employ frozen VGGT~\cite{wang2025vggtvisualgeometrygrounded}, a pretrained model that directly infers 3D geometric attributes including depth maps, point clouds, and camera parameters from visual observations, as a teacher network to guide the learning of geometry-aware features. Critically, rather than adding geometric knowledge along with DINO-SigLIP features into LLM, we align the \textit{LLM's hidden states} corresponding to visual tokens with VGGT's geometric features. This design ensures that geometric understanding is deeply integrated into the multimodal representations that drive action prediction, rather than remaining isolated in the visual processing pipeline. The model is trained with a combined objective that simultaneously optimizes latent action prediction and geometry alignment, enabling it to learn both task-specific visuomotor skills and generalizable geometric reasoning.

We conduct extensive experiments on LIBERO~\cite{liu2023liberobenchmarkingknowledgetransfer} and LIBERO-PRO~\cite{zhou2025liberoprorobustfairevaluation}. On LIBERO, a standard benchmark for language-conditioned manipulation across four task suites, GLaD achieves \textbf{94.1\%} average success rate, outperforming UniVLA (92.5\%) which uses both identical pretraining and posttraining data, and substantially surpassing other strong baselines including OpenVLA (76.5\%), Octo (75.1\%), and Diffusion Policy (72.4\%). Notably, GLaD demonstrates particularly strong performance on object manipulation tasks, achieving \textbf{97.4\%} success rate on LIBERO-OBJECT, the highest among all evaluated methods. On LIBERO-PRO, a robustness benchmark that introduces controlled perturbations across object appearance, spatial layout, language semantics, and task composition to distinguish genuine task understanding from mere memorization, GLaD exhibits substantially improved robustness to visual appearance variations. Under object perturbations that modify color, texture, and size while preserving semantic equivalence, GLaD achieves \textbf{81\%} success rate on LIBERO-GOAL compared to UniVLA's 62\%, and \textbf{54\%} on LIBERO-LONG versus 47\%—with specific tasks showing up to 60 percentage point improvements. These results validate that geometry-aware pretraining enables the model to learn intrinsic geometric features and manipulation affordances rather than relying on superficial visual characteristics, enhancing policy generalization beyond pattern matching.

Our main contributions are as follows:
\begin{itemize}
    \item We identify a critical limitation in current VLA architectures, \emph{i.e.}, the lack of geometric understanding due to reliance on 2D vision encoders (\emph{e.g.}, CLIP, SigLIP) that do not encode spatial positions and object relations. We demonstrate that injecting geometric priors during pretraining can substantially enhance scene understanding and policy generalization.

    \item A geometry-aware VLA framework GLaD is proposed to incorporate 3D geometric knowledge through knowledge distillation. By leveraging VGGT as a frozen teacher network, we distill geometric features into the LLM's hidden states corresponding to visual tokens, ensuring geometric understanding is deeply fused into the multimodal representations that drive action prediction, without requiring depth sensors or explicit 3D annotations.

    \item GLaD achieves an average success rate of 94.1\% on the LIBERO benchmark, surpassing the baseline model UniVLA (92.5\%). Furthermore, on the LIBERO-PRO robustness benchmark, GLaD demonstrates substantially improved resilience to visual appearance variations, achieving 81\% on LIBERO-GOAL under object perturbations ( \emph{vs} 62\% for UniVLA), validating the generalization capability of the proposed geometry-aware pretraining.
\end{itemize}

\section{Related Works}



\noindent \textbf{Vision-Language-Action Models.}
Recent studies have extended large vision-language models (VLMs) to build general-purpose robotic policies capable of generating actions directly from visual and textual inputs.
Early works~\cite{bu2025univlalearningacttaskcentric, brohan2023rt1roboticstransformerrealworld, brohan2023rt2visionlanguageactionmodelstransfer, octomodelteam2024octoopensourcegeneralistrobot, cheang2024gr2generativevideolanguageactionmodel, kim2024openvlaopensourcevisionlanguageactionmodel, li2024cogactfoundationalvisionlanguageactionmodel, black2024pi0visionlanguageactionflowmodel, intelligence2025pi05visionlanguageactionmodelopenworld, intelligence2025pi06vlalearnsexperience, liu2025rdt1bdiffusionfoundationmodel, chi2024diffusionpolicyvisuomotorpolicy} primarily learn from 2D visual observations, relying on implicit reasoning over spatial structures.
Subsequent research began to explicitly incorporate 3D spatial information to enhance spatial understanding and cross-embodiment generalization. For instance, SpatialVLA leverages RGB-D inputs and a Depth API~\cite{qu2025spatialvlaexploringspatialrepresentations}, OG-VLA transforms multi-view RGB-D observations into point clouds and orthographic projections~\cite{singh2025ogvlaorthographicimagegeneration}, PointVLA directly consumes point cloud data~\cite{li2025pointvlainjecting3dworld}, and 4D-VLA extends this idea by integrating temporal sequences of RGB-D inputs~\cite{zhang20254dvlaspatiotemporalvisionlanguageactionpretraining}.
More recent works attempt to implicitly encode 3D geometry without requiring explicit depth sensors: SpatialBot estimates depth using ZoeDepth~\cite{cai2025spatialbotprecisespatialunderstanding, bhat2023zoedepthzeroshottransfercombining}, 3D-VLA learns to infer spatial representations internally~\cite{zhen20243dvla3dvisionlanguageactiongenerative}, and GeoVLA reconstructs 3D embeddings from 2D images through depth estimation and point-cloud generation~\cite{sun2025geovlaempowering3drepresentations}.
Despite these advances, achieving consistent alignment between 3D spatial representations, 2D visual features, and language instructions remains a fundamental challenge in developing unified and robust VLA frameworks.


\noindent \textbf{Geometry-Aware Visual Representation Learning.} A large body of works focus on learning 3D geometry from 2D images. Representative tasks include monocular depth estimation~\cite{chen2025midasmultimodalinteractivedigitalhuman, ranftl2021visiontransformersdenseprediction, bhat2023zoedepthzeroshottransfercombining}, normal prediction~\cite{bae2024dsine, ye2024stablenormal, Bae2021, fan2021three}, and single-view or multi-view reconstruction with implicit 3D representations such as NeRF~\cite{mildenhall2020nerfrepresentingscenesneural} and neural surface fields. These approaches demonstrate that rich geometric priors can be extracted directly from RGB inputs.
Building upon these foundations, recent geometry-grounded vision models~\cite{wang2025vggtvisualgeometrygrounded, wang2025pi3permutationequivariantvisualgeometry, xu20254dgt, xu2025uniugg, huang2025, dust3r_cvpr24} aim to learn latent features that explicitly encode 3D scene structure. Notably, VGGT~\cite{wang2025vggtvisualgeometrygrounded} jointly predicts depth, point clouds, and camera parameters from image sequences, producing geometry-aware representations with strong spatial consistency, while PI3~\cite{wang2025pi3permutationequivariantvisualgeometry} learns permutation-invariant geometric embeddings. Such models provide powerful and generalizable geometric priors, making them suitable teachers for distilling 3D structure into downstream representation learning. However, effectively integrating these geometric priors into vision-language-action models without compromising their generalization capability remains an open challenge.


\noindent \textbf{Knowledge Distillation.}
Before the advent of large language models, knowledge distillation was primarily used as a model compression technique, transferring soft predictions~\cite{hinton2015distillingknowledgeneuralnetwork}, intermediate features~\cite{romero2015fitnetshintsdeepnets, zagoruyko2017payingattentionattentionimproving}, or relational structures~\cite{park2019relationalknowledgedistillation} from a large teacher to a smaller student for efficient deployment.
In the LLM era, distillation has expanded from compressing architectures to transferring capabilities, enabling smaller or specialized models to inherit instruction-following, reasoning, and alignment behaviours from powerful foundation models.
Existing approaches can be grouped into three broad families: (1) generation-based supervision~\cite{xu2024surveyknowledgedistillationlarge}, where teachers provide large-scale labeled or synthesized instruction–response data through labeling~\cite{he2024annollmmakinglargelanguage, wang2024pandalmautomaticevaluationbenchmark, mukherjee2023orcaprogressivelearningcomplex, mitra2023orca2teachingsmall, xu2023baizeopensourcechatmodel, yue2023mammothbuildingmathgeneralist, li2024mixeddistillationhelpssmaller}, expansion~\cite{wang2023selfinstructaligninglanguagemodels, Ranaldi_2024, sun2023principledrivenselfalignmentlanguagemodels, luo2025wizardcoderempoweringcodelarge, luo2025wizardmathempoweringmathematicalreasoning, xu2025wizardlmempoweringlargepretrained, dai2023auggptleveragingchatgpttext}, or curated generation~\cite{ding2023enhancingchatlanguagemodels,gunasekar2023textbooksneed,li2023textbooksneediiphi15,wei2024magicoderempoweringcodegeneration,yu2024wavecoderwidespreadversatileenhancement,ye2022zerogenefficientzeroshotlearning,gao2023selfguidednoisefreedatageneration,bonifacio2022inparsdataaugmentationinformation}; (2) representation-level alignment, which aligns hidden states, output distributions, or preference signals through feature-based~\cite{xu2024surveyknowledgedistillationlarge,timiryasov2023babyllamaknowledgedistillation,gu2025minillmknowledgedistillationlarge,liu2023llmqatdatafreequantizationaware} or feedback-based objectives~\cite{luo2025wizardmathempoweringmathematicalreasoning, bai2022constitutionalaiharmlessnessai, cui2024ultrafeedbackboostinglanguagemodels, gu2025minillmknowledgedistillationlarge, chen-etal-2023-personalized, tunstall2023zephyrdirectdistillationlm, hong2023cyclealigniterativedistillationblackbox, lee2024rlaifvsrlhfscaling, jiang2023lionadversarialdistillationproprietary}; and (3) self-bootstrapped distillation, where models iteratively refine their own generations without a stronger teacher~\cite{wang-etal-2023-self-instruct, yang2024rlcdreinforcementlearningcontrastive, huang2022largelanguagemodelsselfimprove, gulcehre2023reinforcedselftrainingrestlanguage, yuan2025selfrewardinglanguagemodels, zelikman2022starbootstrappingreasoningreasoning}.
Extending these trends to the vision-language-action domain, recent work investigates how to distill geometric priors from pretrained 3D models into multimodal models to overcome the inherent 2D bias of their visual encoders. Spatial Forcing aligns intermediate VLA representations with geometric embeddings to improve spatial precision in robotic control~\cite{li2025spatialforcingimplicitspatial}; Vid-LLM injects reconstruction-derived geometric cues into video-based multimodal LLMs for enhanced 3D scene reasoning~\cite{chen2025vidllmcompactvideobased3d}; and 3D-Aware VLMs with Geometric Distillation transfer sparse correspondences, depth relations, and cost volumes into vision–language models to augment their 3D spatial understanding~\cite{lee20253dawarevisionlanguagemodelsfinetuning}.
\section{Methodology}
\label{sec:method}

We propose GLaD, an end-to-end VLA framework that integrates an LLM backbone, a vision encoder, an action head, and a geometry distillation module to enable the LLM to extract geometric information from images and generate latent actions conditioned on embodied task instructions.
In Section~\ref{subsec:geometry_distillation}, we present our geometry distillation module that enhances the VLA with 3D geometric understanding. In Section~\ref{subsec:training_strategy}, we detail the training strategy.




\begin{figure*}[!t]
\centering

\begin{minipage}{1\textwidth}
  \centering
  \includegraphics[width=0.92\linewidth]{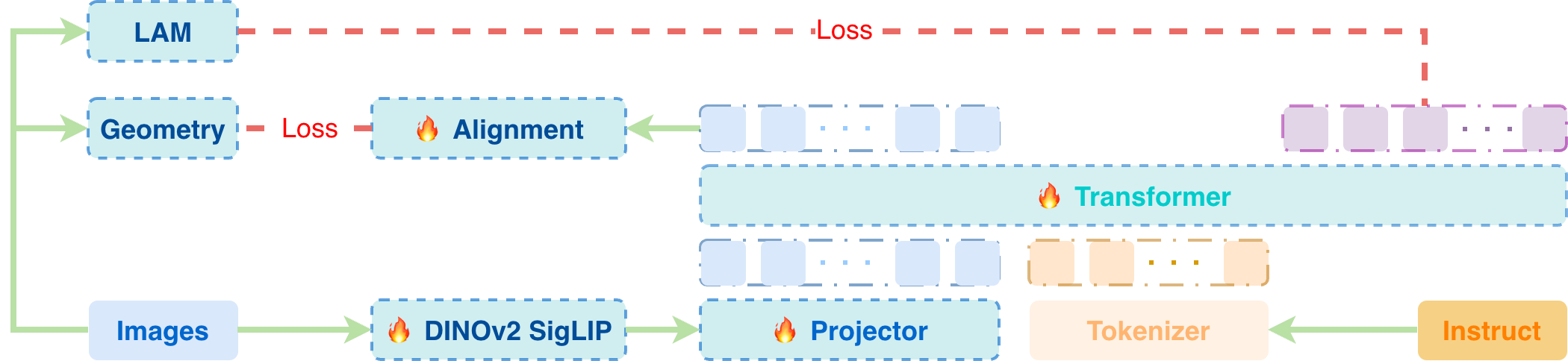}
  \caption*{(a)}
\end{minipage}

\vspace{0.2em}
\makebox[\textwidth]{\dotfill}
\vspace{0.2em}

\begin{minipage}{1\textwidth}
  \centering
  \includegraphics[width=0.92\linewidth]{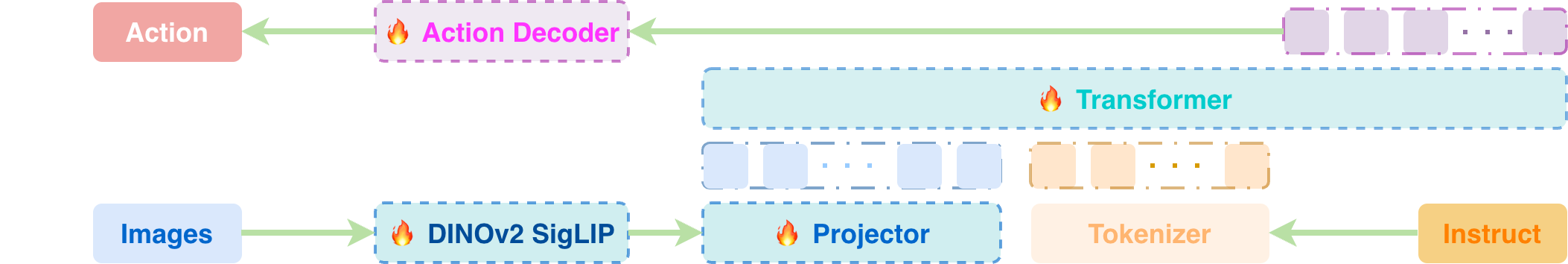}
  \caption*{(b)}
\end{minipage}

\caption{GLaD model architecture. (a) \textbf{Pretraining stage}: The vision encoder (DINOv2~\cite{oquab2024dinov2learningrobustvisual} + SigLIP~\cite{zhai2023sigmoidlosslanguageimage}), projector, and LLM backbone (LLaMA-2-7B~\cite{touvron2023llama2openfoundation}) are trained, while the frozen VGGT teacher provides 3D geometric supervision. The feature alignment module learns to align LLM hidden states with VGGT features. (b) \textbf{Posttraining stage}: The VLA backbone is adapted via LoRA, while the action decoder and feature alignment module are fully trained. The VGGT remains frozen to preserve geometric priors.}
\label{fig:overview}
\end{figure*}

\subsection{Geometry Distillation}
\label{subsec:geometry_distillation}

The overall architecture of GLaD is illustrated in Fig.~\ref{fig:overview}. Our VLA backbone follows the UniVLA architecture~\cite{bu2025univlalearningacttaskcentric}, comprising a Prismatic vision encoder (DINOv2~\cite{oquab2024dinov2learningrobustvisual} and SigLIP~\cite{zhai2023sigmoidlosslanguageimage}), an MLP projector, LLaMA-2-7B backbone~\cite{touvron2023llama2openfoundation}, and an action decoder. To enhance this backbone with 3D geometric understanding, we introduce a geometry distillation module that aligns the LLM's internal visual representations with features from a pretrained geometry-aware teacher network. This module comprises two subcomponents:

\subsubsection{VGGT Feature Extractor}
\label{subsubsec:vggt_extractor}

Following 3DRS, we adopt pretrained VGGT as the teacher network for 3D geometry representation. Given a sequence of historical frames $\{o_{t-T}, ..., o_t\}$ ($T=32$), VGGT produces a spatio-temporal representation $\mathbf{F}_{3d} \in \mathbb{R}^{T \times L \times d_{\text{vggt}}}$ with $d_{\text{vggt}} = 2048$. In GLaD, only a single historical frame is used for simplicity. The VGGT parameters remain frozen throughout the training process.

For the VGGT temporal features $\mathbf{F}_{3d}$, we first apply adaptive pooling to match the spatial dimension with $N_p$ (i.e., the number of visual patches). A ``last-frame'' aggregation strategy is then applied to generate a single-frame representation $\mathbf{F}_{3d}^{\text{single}} \in \mathbb{R}^{N_p \times d_{\text{vggt}}}$.

\subsubsection{Feature Alignment Network}
\label{subsubsec:feature_alignment}

This network projects the final-layer LLM hidden states corresponding to image tokens into the VGGT feature space via a two-layer MLP:
\begin{equation}
\mathbf{H}_{\text{aligned}} = \text{MLP}(\mathbf{H}_{\text{img}}) \in \mathbb{R}^{N_p \times d_{\text{vggt}}},
\end{equation}
where $\mathbf{H}_{\text{img}} \in \mathbb{R}^{N_p \times d_{\text{llm}}}$ denotes the LLM hidden states at image token positions. We extract features from LLM hidden states rather than the vision encoder to ensure geometric knowledge is integrated into fused multimodal representations.

\subsubsection{Training Objective}
\label{subsubsec:integration}

During pretraining, GLaD optimizes a combined loss:
\begin{equation}
\mathcal{L}_{\text{total}} = \mathcal{L}_{\text{VLA}} + \lambda \mathcal{L}_{\text{distill}},
\end{equation}
where $\mathcal{L}_{\text{VLA}}$ is the cross-entropy loss for latent action prediction:
\begin{equation}
\mathcal{L}_{\text{VLA}} = -\sum_{i=1}^{N} \log P(\alpha_i | o, l, \alpha_{<i}),
\end{equation}
and $\mathcal{L}_{\text{distill}}$ is the MSE loss for geometry alignment:
\begin{equation}
\mathcal{L}_{\text{distill}} = \| \mathbf{H}_{\text{aligned}} - \mathbf{F}_{3d}^{\text{single}} \|_2^2.
\end{equation}
The hyperparameter $\lambda$ balances action prediction and geometric alignment. The VGGT teacher remains frozen throughout training.

\subsection{Training Strategy}
\label{subsec:training_strategy}

Our training procedure consists of two stages: large-scale pretraining with geometry distillation and task-specific posttraining.

\subsubsection{Stage 1: Pretraining with Geometry Distillation}

During the pretraining stage (Fig.~\ref{fig:overview}(a)), GLaD is trained on a large-scale robotic manipulation dataset with the combined loss function described in Section~\ref{subsubsec:integration}. The model learns to predict latent actions while simultaneously aligning its internal visual representations with VGGT geometry features.

We initialize the VLA backbone with the pretrained UniVLA checkpoint and introduce the learnable alignment network. The VGGT teacher network remains frozen throughout pretraining. Training is conducted for 45 epochs using AdamW optimizer with a learning rate of 5e-7 on 8$\times$A100 GPUs for approximately 9 days. The distillation loss weight $\lambda$ is tuned based on validation performance to balance action prediction accuracy and geometric alignment.

\subsubsection{Stage 2: Posttraining on Downstream Tasks}

After pretraining, we adapt GLaD to specific downstream tasks (e.g., LIBERO) through supervised fine-tuning (Fig.~\ref{fig:overview}(b)). During this stage, the VLA backbone is adapted via LoRA~\cite{hu2021loralowrankadaptationlarge} for parameter-efficient fine-tuning, while the action decoder and feature alignment module are fully trained. The VGGT teacher network remains frozen to preserve the learned geometric priors. Task-specific posttraining is conducted for 60k steps with learning rate 3.5e-5 on 8$\times$A100 GPUs.

\section{Experiments}

We train and evaluate GLaD across three stages: large-scale pretraining on the Bridge dataset, post-training on LIBERO dataset, and comprehensive evaluation on the standard LIBERO benchmark and enhanced LIBERO-PRO benchmark. In Section~\ref{subsec:data_benchmarks}, we introduce the datasets and benchmarks used in our experiments. In Section~\ref{subsec:training}, we detail our training protocol including pretraining and post-training phases. In Section~\ref{subsec:libero_results}, we present evaluation results on the standard LIBERO benchmark, where GLaD achieves state-of-the-art 94.1\% average success rate. In Section~\ref{subsec:liberopro_results}, we discuss the robustness evaluation framework LIBERO-PRO. Finally, in Section~\ref{subsec:ablation}, we conduct ablation studies to analyze the impact of pretraining checkpoint selection and post-training duration.

\subsection{Datasets and Benchmarks}
\label{subsec:data_benchmarks}

\subsubsection{Bridge Dataset}

We use the Bridge dataset~\cite{ebert2021bridgedataboostinggeneralization} for large-scale pretraining. The Bridge dataset provides diverse manipulation demonstrations that help the model acquire foundational visuomotor skills. We choose Bridge over larger datasets like OXE because Bridge alone provides sufficient diversity and scale for our pretraining objectives, while being more computationally efficient.

\subsubsection{LIBERO Benchmark and LIBERO Dataset}

LIBERO~\cite{liu2023liberobenchmarkingknowledgetransfer} is a benchmark for lifelong learning in robot manipulation, featuring procedurally generated tasks based on everyday human activities. The benchmark includes 130 language-conditioned manipulation tasks organized into four suites, each designed to evaluate different aspects of knowledge transfer:

\textbf{LIBERO-SPATIAL} (10 tasks) tests the transfer of spatial knowledge. All tasks require placing a bowl on a plate among the same set of objects, but the bowl's location varies across tasks. Success requires continually learning and memorizing new spatial relationships.

\textbf{LIBERO-OBJECT} (10 tasks) evaluates object-level knowledge transfer. Each task involves pick-and-place of a unique object, requiring the agent to recognize and manipulate different object types.

\textbf{LIBERO-GOAL} (10 tasks) assesses procedural knowledge transfer. All tasks share the same objects and spatial layout but differ in task goals, requiring the agent to learn diverse manipulation behaviors.

\textbf{LIBERO-LONG} (10 tasks) contains long-horizon manipulation tasks that combine multiple subtasks, testing the model's ability to handle complex, multi-step procedures.

Each task is accompanied by 50 high-quality human teleoperation demonstrations. Following LIBERO protocol, we evaluate on 50 episodes per task.

\subsubsection{LIBERO-PRO Benchmark}

While LIBERO provides a standardized evaluation framework, recent work~\cite{zhou2025liberoprorobustfairevaluation} has revealed critical limitations: models achieving over 90\% success on standard LIBERO often fail completely under minor perturbations, suggesting reliance on memorization rather than genuine task understanding.

To test this, we evaluate on LIBERO-PRO~\cite{zhou2025liberoprorobustfairevaluation}, which extends LIBERO with controlled perturbations across four dimensions:

\textbf{Object Perturbations} modify non-essential object attributes (color, texture, size) while preserving semantic equivalence, testing robustness to superficial visual changes.

\textbf{Position Perturbations} alter initial object placements, both absolute positions and relative spatial arrangements, probing spatial reasoning under varied layouts.

\textbf{Semantic Perturbations} rephrase task instructions while preserving the original task intent (e.g., ``pick up'' $\rightarrow$ ``grab'', ``place on'' $\rightarrow$ ``put on top of''), evaluating whether the model genuinely understands language semantics or merely pattern-matches specific phrasings.

\textbf{Task Perturbations} modify the task itself by changing target objects or required actions, while ensuring all components (objects and actions) appear in the training set. This tests compositional generalization—the ability to recombine known elements in novel ways.

Unlike the standard LIBERO evaluation where test tasks closely mirror training tasks, LIBERO-PRO introduces sufficient variation to distinguish between memorization and genuine generalization. And for LIBERO-PRO, we focus our evaluation on the perturbation types that are consistently available across all task suites in the current benchmark release.

\subsection{Training Protocol}
\label{subsec:training}

\subsubsection{Pretraining Phase}

We pretrain the model on the Bridge dataset using 8$\times$A100 GPUs for 9 days, spanning 45 epochs with learning rate 5e-7. Pretraining enables the model to acquire general visuomotor skills from large-scale diverse data before specializing on LIBERO tasks. We use checkpoints at epochs 27 and 45, both of which serve as initialization for subsequent post-training experiments.

\subsubsection{Post-training Phase}

We perform task-specific post-training on the LIBERO dataset to adapt the model to LIBERO's task distribution. We train for 48k steps with learning rate 3.5e-5, saving checkpoints every 16k steps. All post-training experiments use 8$\times$A100 GPUs. We select the best-performing checkpoints based on validation performance for final evaluation.

\subsection{Evaluation on LIBERO Benchmark}
\label{subsec:libero_results}

\begin{table}[!t]
\caption{\label{tab:libero}\textbf{Results on LIBERO benchmark across four evaluation suites.} We compare GLaD against state-of-the-art VLA baselines across spatial reasoning (LIBERO-SPATIAL), object manipulation (LIBERO-OBJECT), goal-oriented tasks (LIBERO-GOAL), and long-horizon procedures (LIBERO-LONG). Success rates (\%) are averaged over 50 episodes per task. GLaD achieves competitive performance with 94.1\% average success rate, ranking among top-tier methods and demonstrating particularly strong object manipulation capabilities (97.4\%, highest among all methods). \textbf{Bold} indicates highest performance and {\ul underline} indicates second-highest performance.}
\centering
\begin{tabular}{@{}cccccc@{}}
\toprule
                   & spatial & object & goal & long & average \\ \midrule
lapa               & 73.8          & 74.6          & 58.8          & 55.4          & 65.7          \\
diffusion   policy & 78.3          & 92.5          & 68.3          & 50.5          & 72.4          \\
octo               & 78.9          & 85.7          & 84.6          & 51.1          & 75.1          \\
mdt                & 78.5          & 87.5          & 73.5          & 64.8          & 76.1          \\
openvla            & 84.7          & 88.4          & 79.2          & 53.7          & 76.5          \\
mail               & 74.3          & 90.1          & 81.8          & 78.6          & 81.2          \\ 
univla      & \textbf{95.2} & {\ul 95.4}    & {\ul 91.9}    & {\ul 87.5}    & {\ul 92.5}    \\
GLaD                & {\ul 95}      & \textbf{97.4} & \textbf{94.4} & \textbf{89.4} & \textbf{94.1} \\ \bottomrule
\end{tabular}
\end{table}

\textbf{Evaluation Setup:} We evaluate on all four LIBERO task suites using the standard simulator environment. Following the LIBERO protocol, we report success rates averaged over 50 evaluation episodes per task. We use the data processing pipeline from OpenVLA to exclude failure demonstrations during training.

\textbf{Results:} Table~\ref{tab:libero} summarizes the performance across all task suites. Our GLaD, pretrained only on Bridge dataset, achieves 94.1\% average success rate, outperforming UniVLA (92.5\%) which uses the same pretraining data. This demonstrates that geometry-aware pretraining provides efficiency gains comparable to data scaling. GLaD also substantially outperforms other baselines including MAIL, OpenVLA, MDT, Octo , and Diffusion Policy.

GLaD demonstrates strong performance across three suites: LIBERO-OBJECT (97.4\%), LIBERO-GOAL (94.4\%), and LIBERO-LONG (89.4\%). The particularly strong performance on LIBERO-OBJECT validates that geometry-aware pretraining effectively captures object-level visual features and manipulation affordances. The consistent improvements over UniVLA across all suites demonstrate the effectiveness of incorporating geometric structure into VLA pretraining.

\subsection{Robustness Analysis on LIBERO-PRO}
\label{subsec:liberopro_results}

\begin{figure}[!t]
\centering
\includegraphics[width=\linewidth]{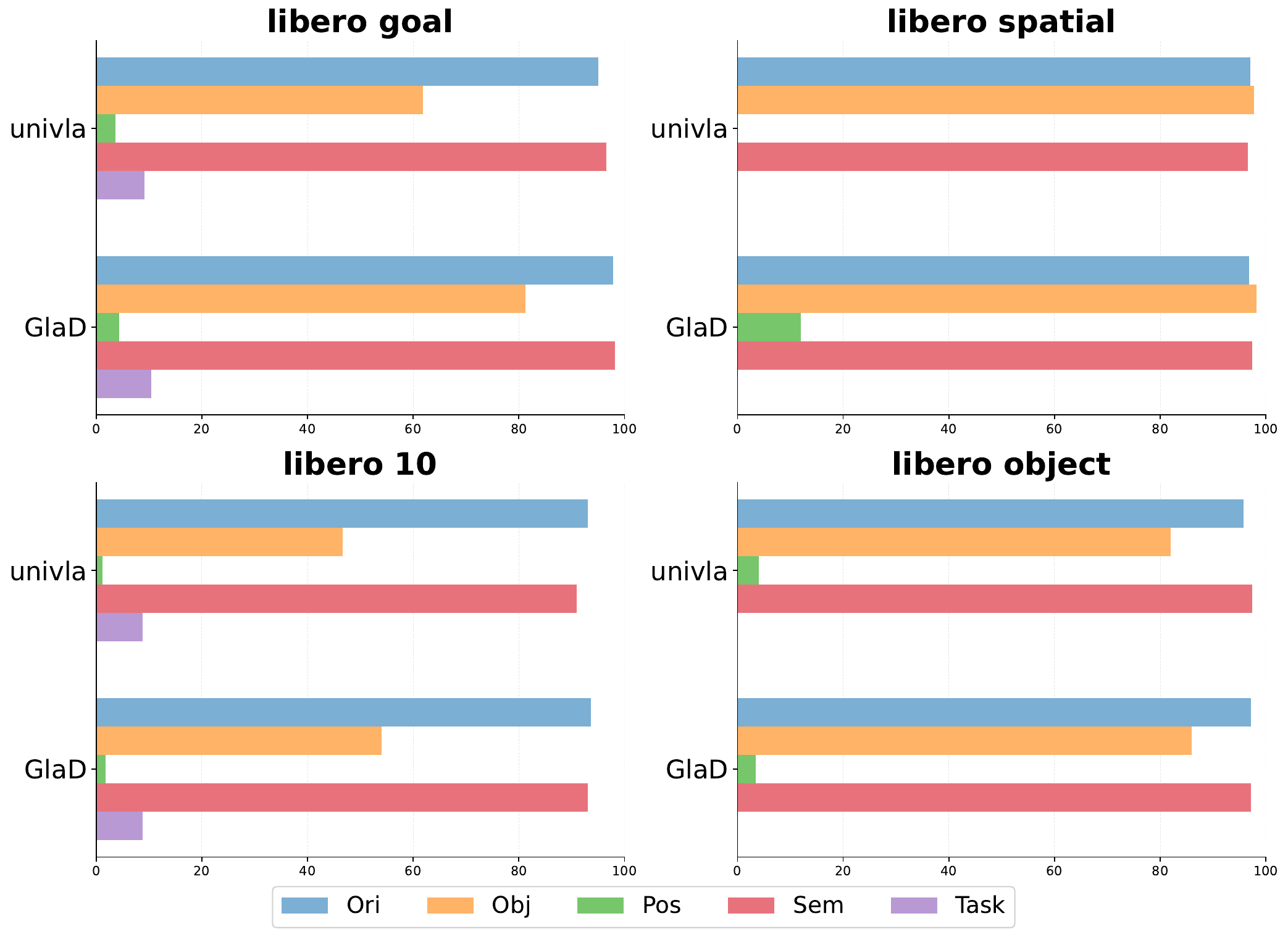}
\caption{\label{fig:combined_chart}\textbf{Robustness comparison across LIBERO suites under five perturbation types.} We compare GLaD against UniVLA on four LIBERO suites: GOAL, SPATIAL, Long-horizon (10), and OBJECT. \textbf{Ori}: Original tasks; \textbf{Obj}: Object perturbations (color, texture, size); \textbf{Pos}: Position perturbations; \textbf{Sem}: Semantic perturbations (language); \textbf{Task}: Task perturbations. Success rates (\%) averaged over 50 episodes per task. GLaD demonstrates significant improvements in object perturbation robustness, particularly on GOAL (81\% vs 62\%) and Long (54\% vs 47\%).}
\end{figure}

%

\begin{table*}[!t]
\centering
\caption{Average success rates across four LIBERO suites under different perturbation types. The columns represent: \textbf{Ori} (original task without perturbations), \textbf{Obj} (object appearance perturbations including color, texture, and size changes), \textbf{Pos} (position perturbations with spatial layout variations), \textbf{Sem} (semantic perturbations with language rephrasing), and \textbf{Task} (task perturbations with compositional changes using known elements). Results are averaged over all tasks in each suite (50 episodes per task). GLaD demonstrates significant advantages under object perturbations: 81\% on LIBERO-Goal compared to UniVLA's 62\%, and 54\% on LIBERO-10 compared to 47\%, showing improved robustness to visual appearance variations. Complete per-task results are provided in Appendix~\ref{app:libero_pro_detailed}.}
\label{tab:libero_pro_summary}
\begin{tabularx}{0.8\textwidth}{lXXXXXXXXXX}
\toprule
\multirow{2}{*}{Benchmark} & \multicolumn{5}{c}{univla} & \multicolumn{5}{c}{GlaD} \\
\cmidrule(lr){2-6} \cmidrule(lr){7-11}
 & Ori & Obj & Pos & Sem & Task & Ori & Obj & Pos & Sem & Task \\
\midrule
LIBERO-Goal    & 95 & 62 & 4 & 97 & 9 & \textbf{98} & \textbf{81} & 4 & \textbf{98} & \textbf{10} \\
LIBERO-Spatial & 97 & 98 & 0 & 97 & -- & 97 & 98 & \textbf{12} & 97 & -- \\
LIBERO-10      & 93 & 47 & 1 & 91 & 9 & \textbf{94} & \textbf{54} & \textbf{2} & \textbf{93} & 9 \\
LIBERO-Object  & 96 & 82 & 4 & 97 & 0 & \textbf{97} & \textbf{86} & 3 & 97 & 0 \\
\bottomrule
\end{tabularx}

\end{table*}

\textbf{Evaluation Setup:} We evaluate GLaD and UniVLA on LIBERO-PRO to assess robustness under controlled perturbations. As introduced in Section~\ref{subsec:data_benchmarks}, LIBERO-PRO distinguishes between genuine task understanding and memorization through systematic variations in objects, positions, semantics, and task compositions. Figure~\ref{fig:combined_chart} visualizes the overall robustness comparison across all perturbation types, while Table~\ref{tab:libero_pro_summary} summarizes the averaged results across all four suites. Detailed per-task results are provided in Appendix~\ref{app:libero_pro_detailed}.

\textbf{Results:} As shown in Figure~\ref{fig:combined_chart} and Table~\ref{tab:libero_pro_summary}, compared with UniVLA, GLaD demonstrates significant advantages under object perturbations, which modify non-essential visual attributes (color, texture, size) while preserving semantic equivalence. On LIBERO-GOAL, GLaD achieves 81\% average success rate compared to UniVLA's 62\%, a substantial +19 percentage point improvement. This gap is even more pronounced in specific tasks: for ``Put(bowl, plate)'', GLaD reaches 84\% while UniVLA achieves only 24\%—a 60 percentage point difference (detailed results in Table~\ref{tab:libero_goal_detailed}). On LIBERO-LONG, GLaD achieves 54\% compared to UniVLA's 47\% (Table~\ref{tab:libero_10_detailed}), showing improved robustness on long-horizon tasks with appearance variations. On LIBERO-OBJECT, GLaD maintains 86\% success rate versus UniVLA's 82\% (Table~\ref{tab:libero_object_detailed}).

These results validate GLaD's core design principle: geometry-aware pretraining enables the model to learn intrinsic geometric features and manipulation affordances rather than relying on superficial visual characteristics. This proves critical when object appearances change while geometric structure remains constant. The consistent improvements across all four suites demonstrate that geometric understanding generalizes across different task types and complexities.

Both models exhibit strong semantic robustness, achieving 93-98\% success rates across all suites under language rephrasing. This demonstrates that VLA architectures with large language model backbones effectively generalize across language variations, understanding task intent despite different phrasings.

On position perturbations, both models show limited robustness. GLaD achieves 12\% on LIBERO-SPATIAL compared to UniVLA's 0\% (Table~\ref{tab:libero_spatial_detailed}), suggesting modest improvements in spatial reasoning. However, performance remains low on other suites (1-4\%), indicating that spatial layout variations bring substantial challenges. Task perturbations, which test compositional generalization by recombining known elements in novel ways, remain challenging for both approaches with ~9-10\% success rates. This reveals shared limitations of current VLA methods in handling compositional reasoning and novel task configurations.

Overall, the LIBERO-PRO evaluation demonstrates that GLaD's geometry-aware pretraining provides substantial robustness advantages specifically in scenarios involving visual appearance variations—precisely the domain where geometric understanding matters most. These findings align with and reinforce the results from standard LIBERO benchmark (Section~\ref{subsec:libero_results}), further validating the effectiveness of incorporating geometric structure into vision-language-action models.

\subsection{Ablation Studies}
\label{subsec:ablation}

We conduct comprehensive ablation experiments to validate the key design choices in GLaD's geometry-aware architecture. Table~\ref{tab:ablation} presents results across three critical dimensions: geometry encoder selection, feature alignment strategy, and geometry integration method. We also provide attention pattern analysis to offer qualitative insights into how these design choices affect task-relevant object localization and manipulation reasoning.

\begin{table}[!t]
\caption{\label{tab:ablation}\textbf{Ablation study on key architectural design choices.} We evaluate three critical design dimensions: geometry encoder selection (VGGT vs. PI3), feature alignment layer (layer 32 vs. layer 24), and geometry integration strategy (late fusion vs. early weighted fusion). Success rates (\%) are averaged over 50 episodes per task across all four LIBERO suites. Bold numbers indicate the best performance in each column. \textbf{GLaD (full)}: VGGT encoder + Layer 32/32 alignment + Late fusion in LLM representation space. \textbf{PI3}: Replaces VGGT with Pi3 encoder (Permutation-Equivariant Visual Geometry Learning encoder). \textbf{Layer 24/32}: Aligns geometric features with layer 24 instead of final layer 32. \textbf{Weighted Fusion}: Aligns VGGT features with DinoSigLIP features, then performs weighted combination before LLM input (early fusion).}
\centering
\begin{tabular}{@{}lccccc@{}}
\toprule
\textbf{Configuration}     & \textbf{Spatial} & \textbf{Object} & \textbf{Goal} & \textbf{Long} & \textbf{Average} \\
\midrule
\textbf{GLaD}              & \textbf{95.0}    & 97.4            & \textbf{94.4} & 89.4          & \textbf{94.1}    \\
\midrule
\multicolumn{6}{l}{\textit{Geometry Encoder Ablation}} \\
\quad PI3                  & 65.2             & \textbf{98.6}   & 94.2          & 86.4          & 86.1             \\
\multicolumn{6}{l}{\textit{Feature Alignment Layer Ablation}} \\
\quad Layer 24/32          & 94.4             & 90.6            & \textbf{94.4} & \textbf{91.0} & 92.6             \\
\multicolumn{6}{l}{\textit{Geometry Integration Strategy Ablation}} \\
\quad Weighted Fusion      & 87.6             & 80.8            & 91.4          & 76.0          & 84.0             \\
\bottomrule
\end{tabular}
\end{table}

\begin{figure*}[t]
\centering
\begin{minipage}{0.95\linewidth}  
  \centering

  \begin{minipage}{0.18\linewidth}
    \centering
    \includegraphics[width=\linewidth]{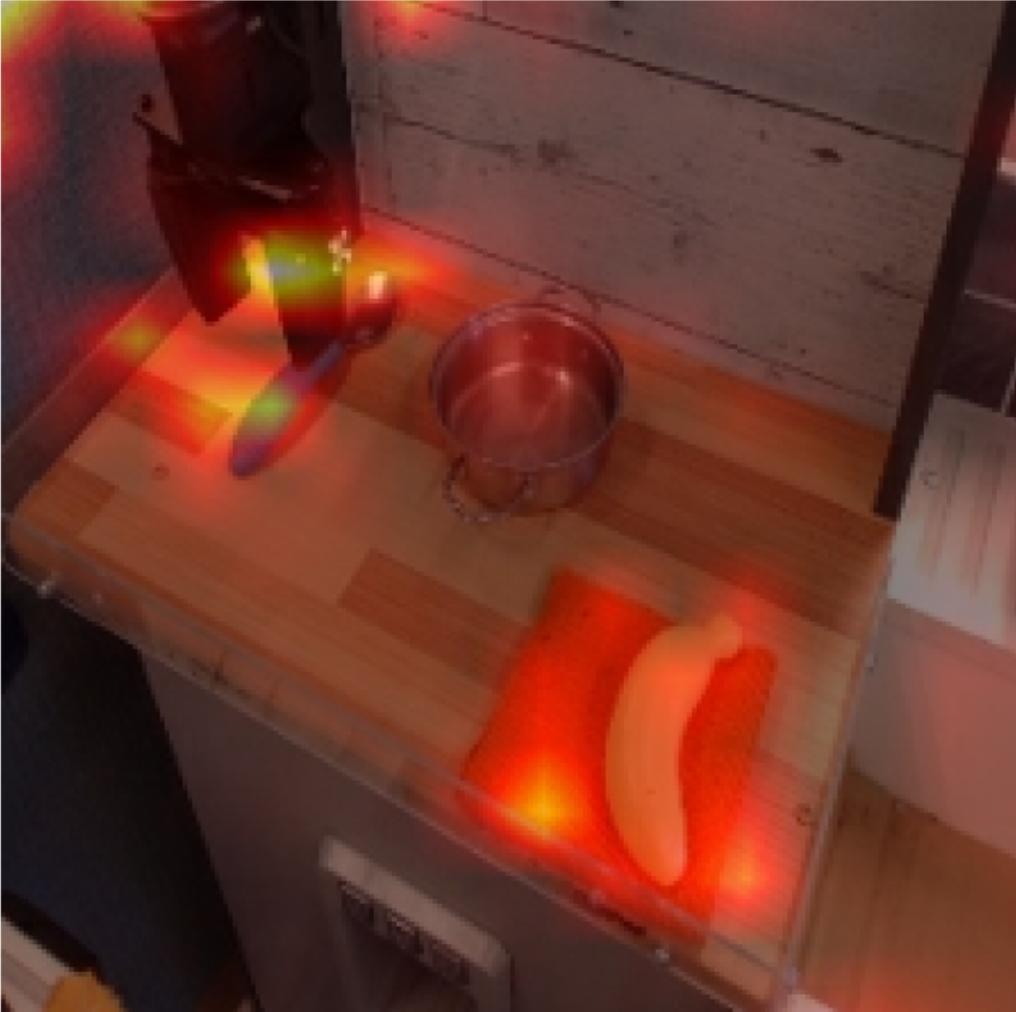}
  \end{minipage}
  \hfill
  \begin{minipage}{0.18\linewidth}
    \centering
    \includegraphics[width=\linewidth]{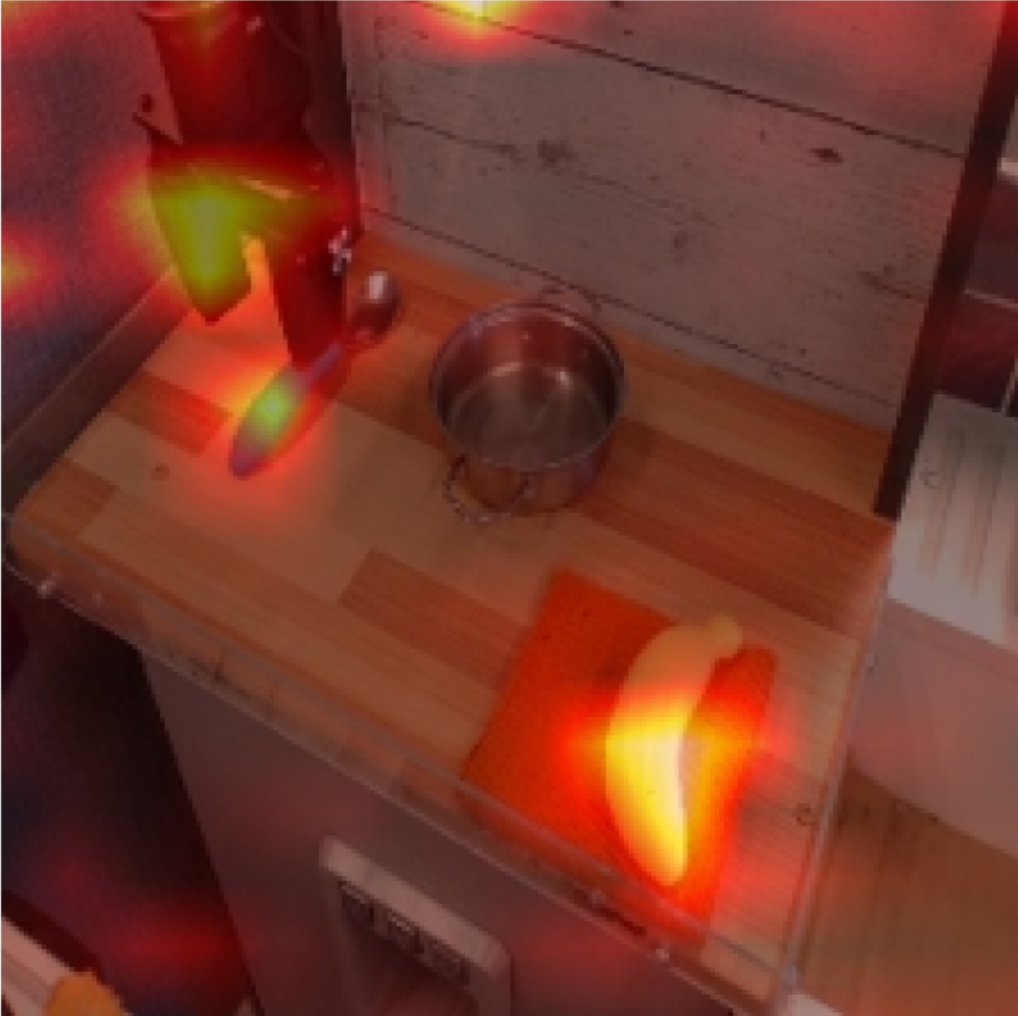}
  \end{minipage}
  \hfill
  \begin{minipage}{0.18\linewidth}
    \centering
    \includegraphics[width=\linewidth]{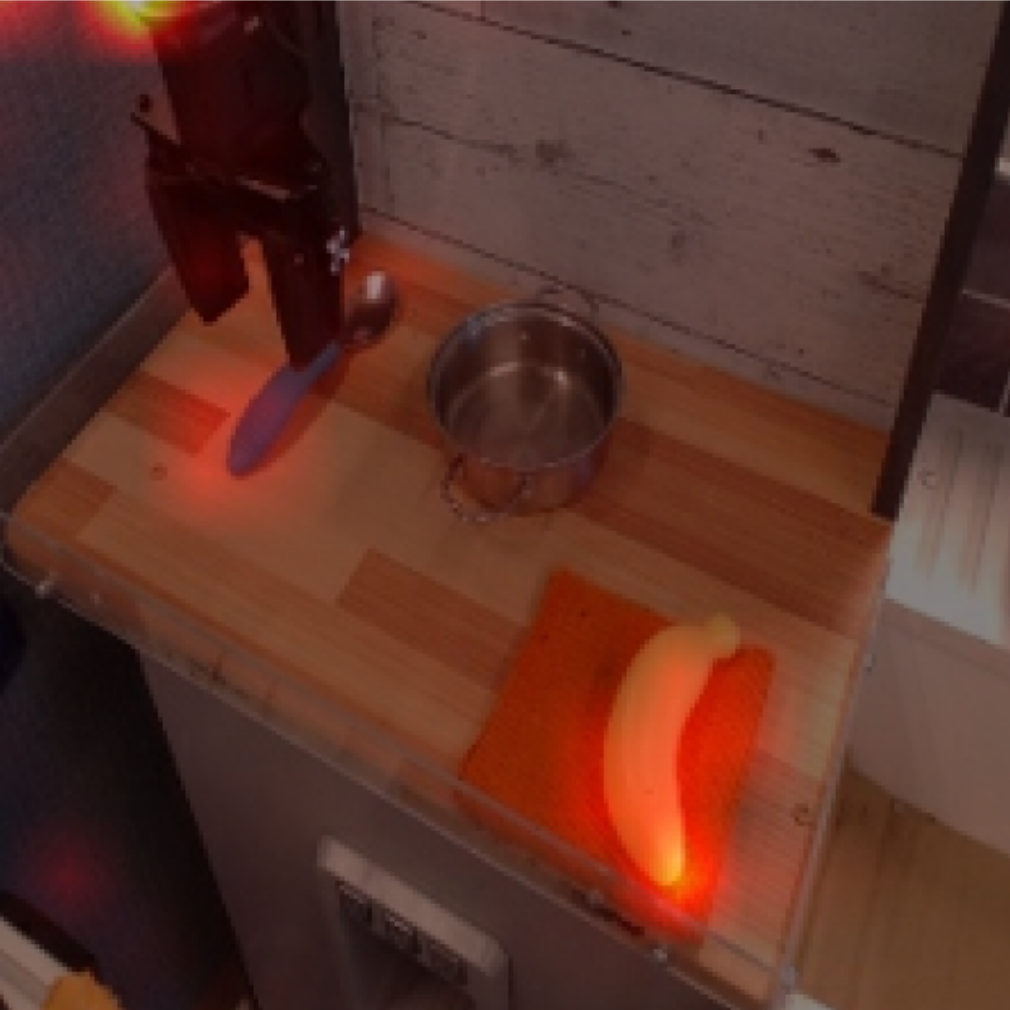}
  \end{minipage}
  \hfill
  \begin{minipage}{0.18\linewidth}
    \centering
    \includegraphics[width=\linewidth]{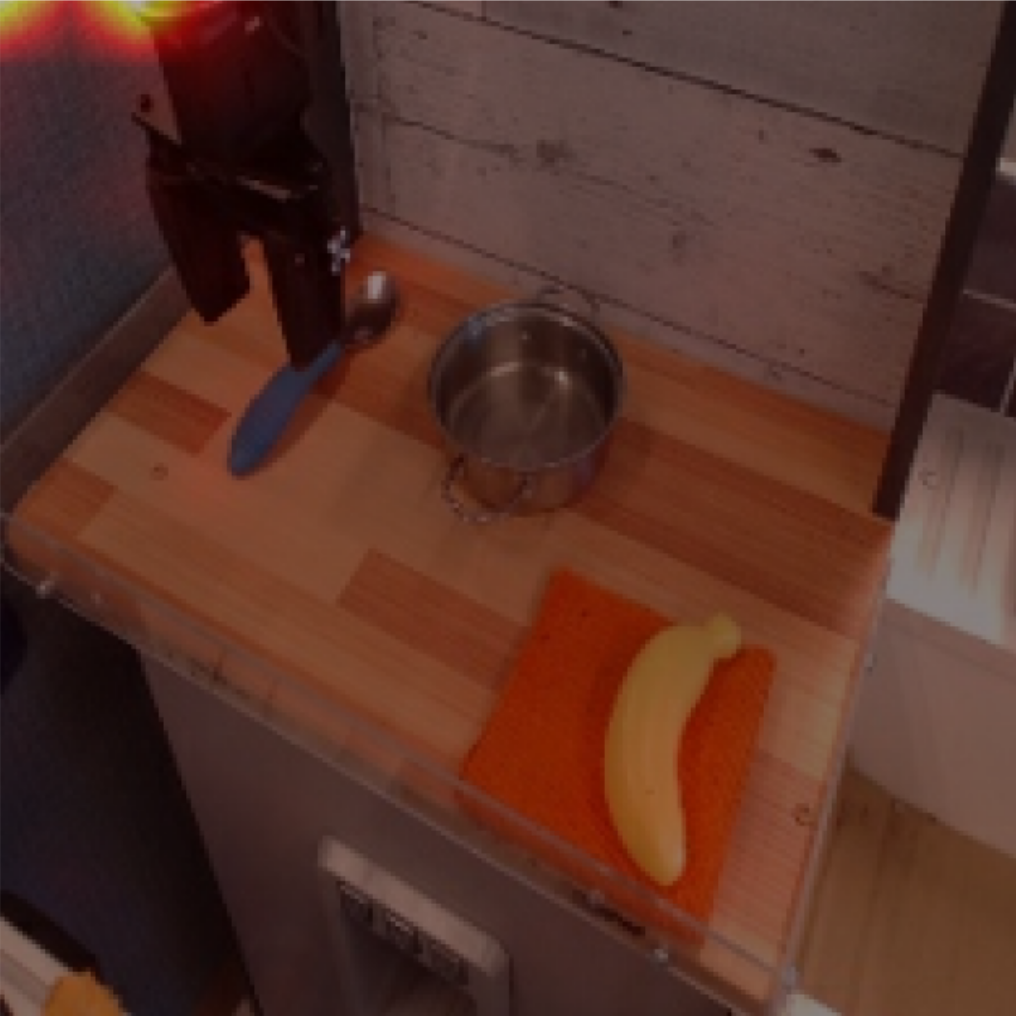}
  \end{minipage}
  \hfill
  \begin{minipage}{0.18\linewidth}
    \centering
    \includegraphics[width=\linewidth]{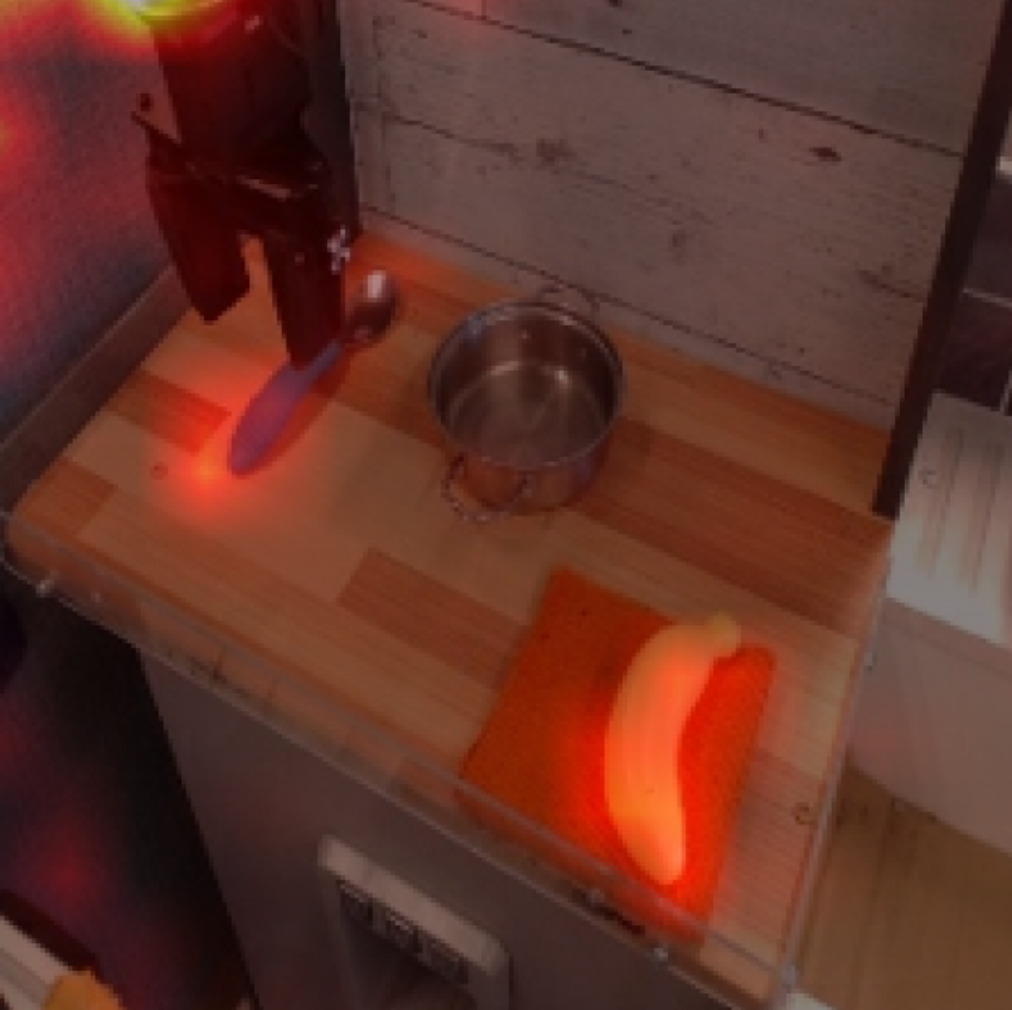}
  \end{minipage}

  \vspace{0.7em}

  \begin{minipage}{0.18\linewidth}
    \centering
    \includegraphics[width=\linewidth]{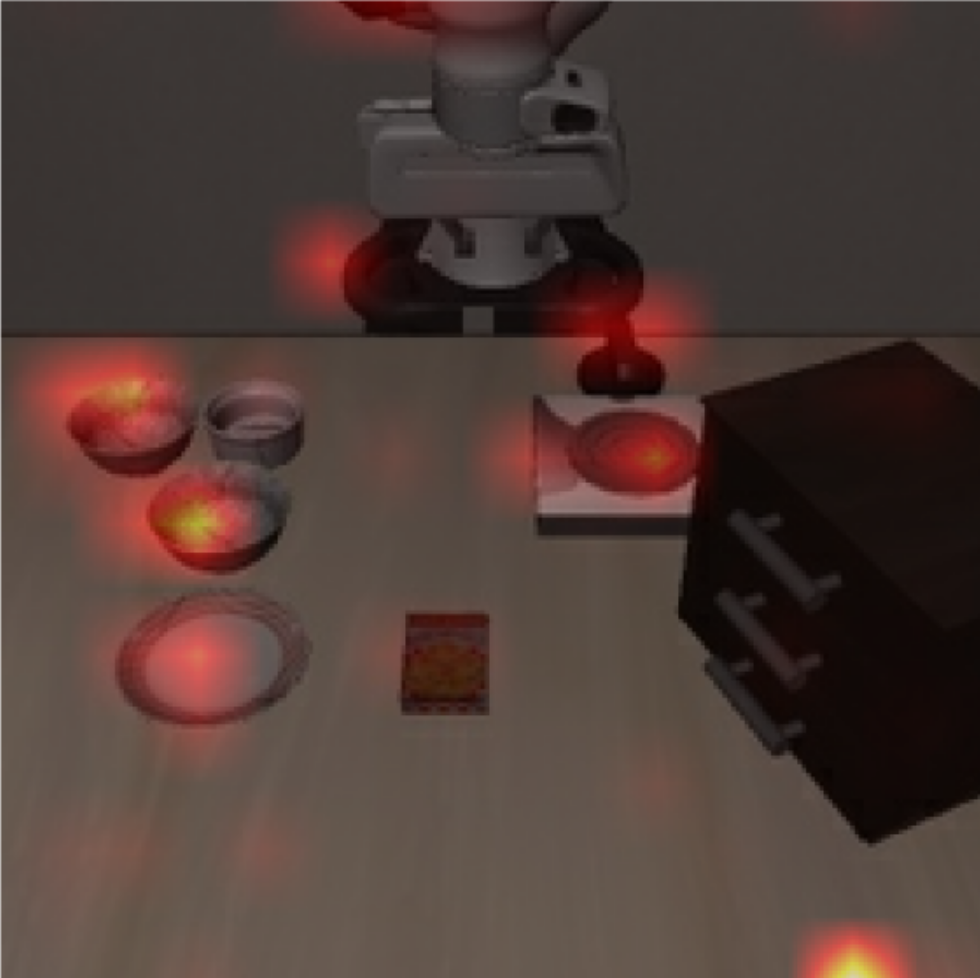}
    \caption*{UniVLA}
  \end{minipage}
  \hfill
  \begin{minipage}{0.18\linewidth}
    \centering
    \includegraphics[width=\linewidth]{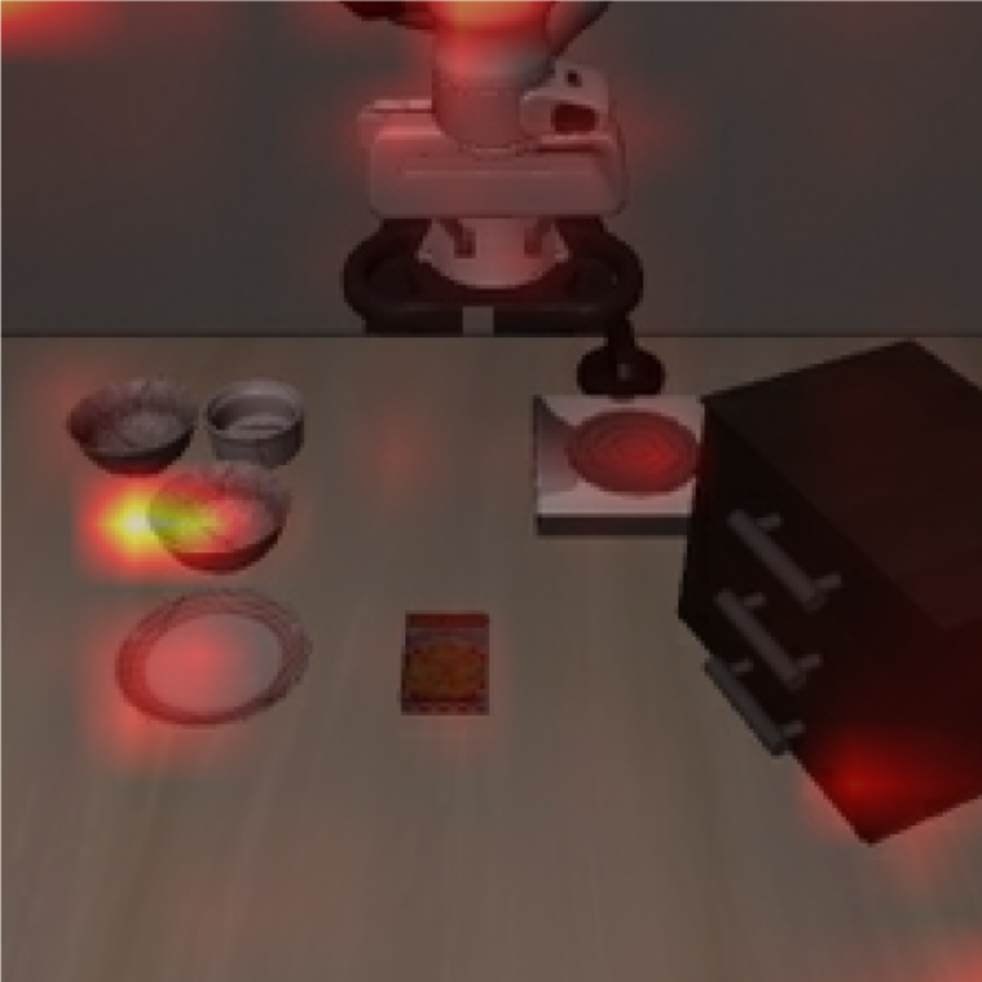}
    \caption*{GLaD}
  \end{minipage}
  \hfill
  \begin{minipage}{0.18\linewidth}
    \centering
    \includegraphics[width=\linewidth]{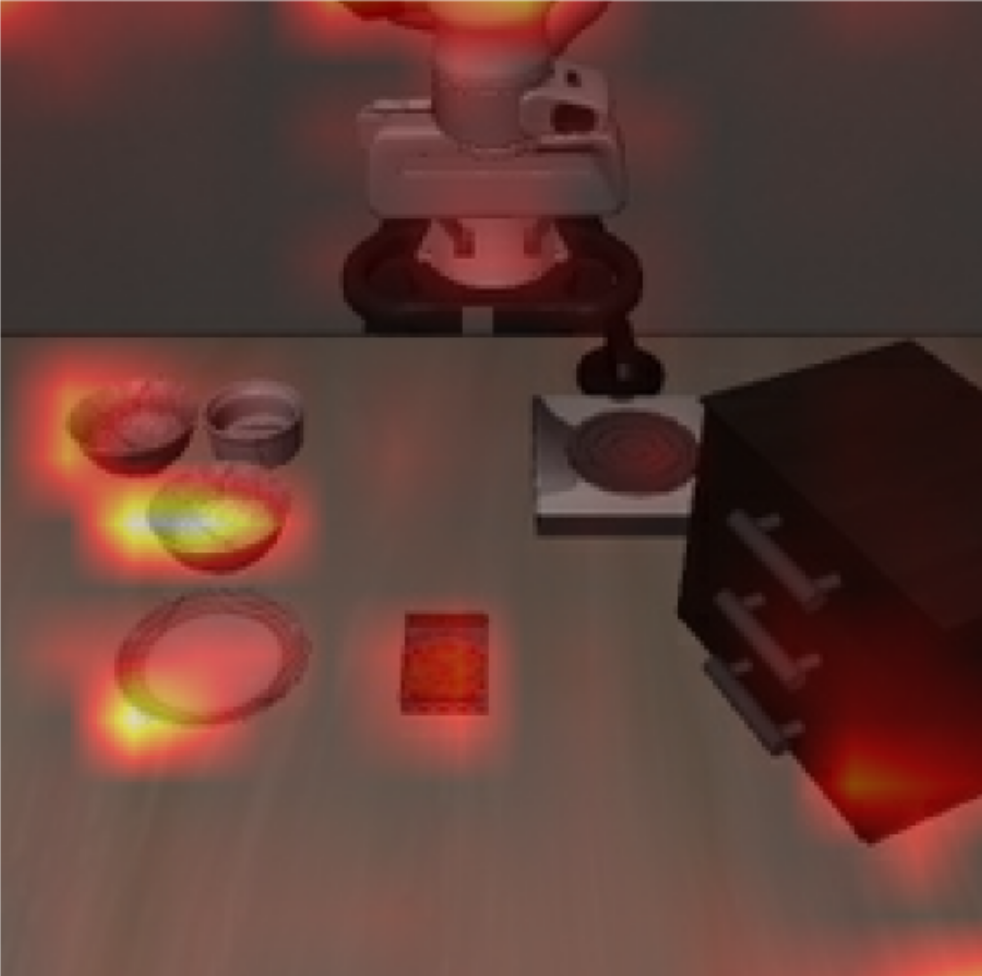}
    \caption*{GLaD-Pi3}
  \end{minipage}
  \hfill
  \begin{minipage}{0.18\linewidth}
    \centering
    \includegraphics[width=\linewidth]{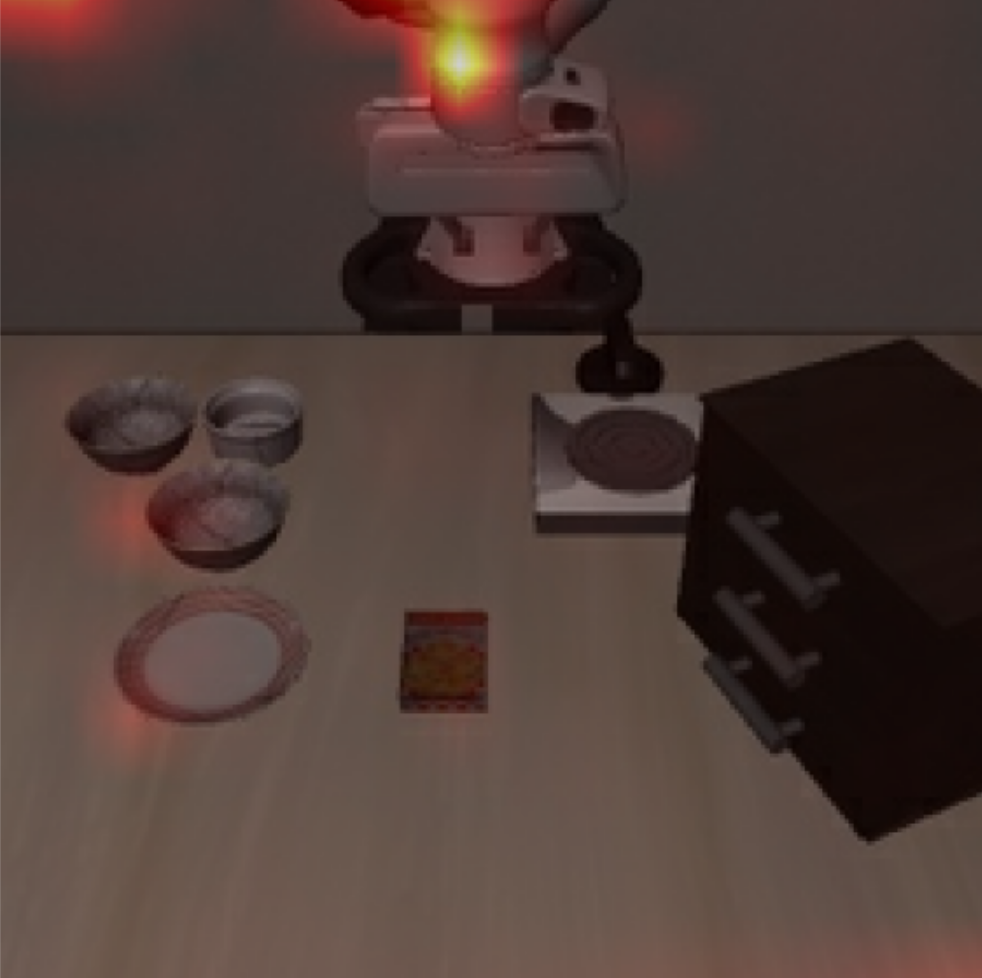}
    \caption*{GLaD-L24}
  \end{minipage}
  \hfill
  \begin{minipage}{0.18\linewidth}
    \centering
    \includegraphics[width=\linewidth]{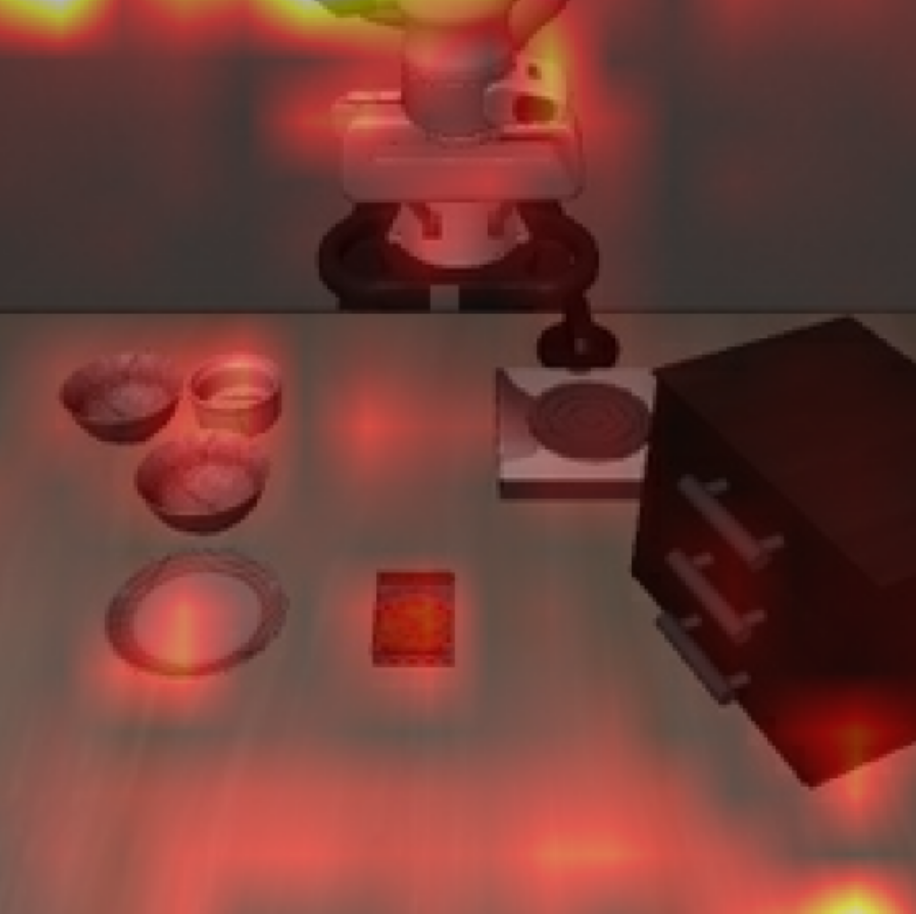}
    \caption*{GLaD-Weighted}
  \end{minipage}

\end{minipage}

\caption{Attention maps across model variants. Up: Bridge scene (``Put the banana in front of the spoon''). Down: LIBERO scene (``Pick up the black bowl between the plate and the ramekin and place it on the plate''). GLaD-Pi3: w/ Pi3 encoder; GLaD-L24: w/ Layer-24 alignment; GLaD-Weighted: w/ early weighted fusion. See Table~\ref{tab:ablation} for detailed configurations and quantitative results.}
\label{fig:ablation_attention_map}
\end{figure*}

\textbf{Geometry Encoder Architecture:} We compare two geometry encoders: VGGT~\cite{wang2025vggtvisualgeometrygrounded} and PI3~\cite{wang2025pi3permutationequivariantvisualgeometry}. Notably, LIBERO-SPATIAL exhibits the highest sensitivity to geometry encoder selection across all ablations, with VGGT achieving 95.0\% compared to 65.2\% with PI3—a 29.8 percentage point difference that represents the largest performance gap among all task suites. This directly validates that VGGT's geometry-grounded visual representation is particularly effective for spatial reasoning tasks, precisely the capability LIBERO-SPATIAL is designed to test. While both encoders achieve strong performance on object-centric tasks (LIBERO-OBJECT: 97.4\% vs. 98.6\%), VGGT's substantial advantage on spatial tasks (94.1\% average vs. 86.1\%) demonstrates that its geometry-aware features provide robust spatial understanding crucial for manipulation tasks requiring precise spatial relationships, such as ``place bowl on plate at specific location''.

\textbf{Feature Alignment Layer:} We investigate the impact of aligning geometric features to different layers of the LLM backbone. Our default configuration aligns to the final layer (32/32), while the ablation variant aligns to an earlier layer (24/32). Early-layer alignment (layer 24) achieves 92.6\% average success rate, showing notable performance drop on LIBERO-OBJECT (90.6\% vs. 97.4\%). This demonstrates that aligning geometric features to the final representation layer is crucial for effective multimodal fusion. Late-layer alignment allows the language model to first process visual-semantic features through most of its depth before integrating geometric information, enabling better preservation of both visual semantics and spatial structure. Early alignment may cause geometric signals to be diluted as they propagate through subsequent transformer layers.

\textbf{Geometry Integration Strategy:} We compare two approaches for incorporating geometric information: (1) our default method that aligns geometry features to the final-layer visual tokens in the LLM's representation space, and (2) an alternative approach that aligns VGGT features to DinoSigLIP features before LLM input, then performs weighted combination. The weighted-feature fusion approach achieves only 84.0\% average success rate with particularly poor performance on LIBERO-OBJECT (80.8\%) and LIBERO-LONG (76.0\%). This substantial degradation suggests that early fusion in the visual feature space, before language model processing, fails to leverage the LLM's capacity for multimodal reasoning. Our late-fusion approach enables the language model to learn task-adaptive integration of geometric and semantic cues, rather than relying on fixed weighted combination.

\textbf{Attention Pattern Analysis:} To provide qualitative insights into the quantitative results above, Figure~\ref{fig:ablation_attention_map} visualizes attention distributions across model variants, revealing how design choices affect task-relevant object localization. GLaD demonstrates sharp, focused attention on manipulation targets (banana in Bridge scene, target plate in LIBERO scene), correlating with its strong performance (94.1\% average). In contrast, GLaD-Pi3 exhibits scattered attention across multiple plates in LIBERO, directly explaining its LIBERO-SPATIAL failure (65.2\%); GLaD-L24 shows diffused attention unable to identify task-relevant regions, aligning with its LIBERO-OBJECT degradation (90.6\%); UniVLA and GLaD-Weighted attend more to the gripper than target objects, indicating reliance on egocentric visual cues rather than object-centric reasoning, which explains GLaD-Weighted's poor performance (84.0\%). These attention patterns provide qualitative evidence supporting Table~\ref{tab:ablation}'s quantitative results.

We validate that GLaD's design choices (VGGT for geometry encoding, final-layer alignment, and late-stage feature integration) work synergistically to achieve strong performance across diverse manipulation scenarios. The 8-10\% performance gaps observed in ablations highlight the importance of each component, particularly for spatial reasoning and object manipulation tasks.

\section{Discussion}

\subsection{Why Geometric Understanding Matters}

Analysis of attention maps (Fig.~\ref{fig:ablation_attention_map}) reveals that GLaD develops sharper attention on manipulation-relevant objects compared to baselines. VLA models trained on 2D vision encoders (CLIP, SigLIP) learn semantic correspondences but struggle to ground semantics in 3D spatial structure. By aligning VGGT geometric features to LLM hidden states, GLaD learns representations capturing both \textit{what} objects are and \textit{what} they look like, proving particularly valuable for object-centric tasks (97.4\% on LIBERO-OBJECT).

\subsection{Design Choices and Robustness}

Our ablation studies (Table~\ref{tab:ablation}) show that late-stage alignment to LLM hidden states substantially outperforms early fusion (94.1\% vs. 84.0\%), enabling task-adaptive integration of geometric and semantic cues. LIBERO-PRO evaluation reveals an asymmetry: GLaD demonstrates strong robustness to object appearance perturbations but limited improvement on position perturbations. This validates our hypothesis—geometric features ground representations in spatial structure rather than superficial appearance, making the model robust when colors or textures change while geometric affordances remain constant.

\subsection{Alternative Approaches and Limitations}

We explored explicit geometry supervision (predicting depth maps) and implicit supervision (contrastive learning), but both failed: explicit supervision caused training divergence due to conflicting objectives, while implicit supervision did not outperform baselines. These failures validate our design of aligning pretrained geometry encoder features to LLM hidden states. Limitations remain in position perturbation robustness, though consistent improvements across LIBERO suites suggest geometric understanding is a valuable inductive bias for VLA models. 
\section{Conclusion}

Current vision-language-action models lack geometric understanding due to reliance on 2D vision encoders (CLIP, SigLIP) that do not encode spatial positions and object relations. We proposed GLaD, a geometry-aware VLA framework that incorporates 3D geometric priors during pretraining through knowledge distillation from a frozen Visual Geometry Grounded Transformer (VGGT). Our key contribution is a late-stage feature alignment mechanism that distills geometric features into the LLM's hidden states corresponding to visual tokens, enabling task-adaptive integration of geometric and semantic cues.

GLaD achieves 94.1\% average success rate on LIBERO benchmark, outperforming UniVLA (92.5\%) trained on identical data. On LIBERO-PRO robustness benchmark, GLaD demonstrates improved resilience to perturbations, particularly excelling under object appearance variations, validating that geometry-aware pretraining enhances policy generalization beyond superficial pattern matching. Ablation studies confirm that VGGT geometry encoding, final-layer alignment, and late-stage integration each contribute significantly to performance.

While limitations remain in spatial layout generalization, our results establish that incorporating geometric priors during pretraining is a promising direction for building more capable vision-language-action models for robotic manipulation.

{
    \small
    \bibliographystyle{IEEEtran}
    \bibliography{main}
}

\newpage

\appendix

\subsection{Detailed LIBERO-PRO Results}
\label{app:libero_pro_detailed}

This appendix provides complete per-task results for the LIBERO-PRO benchmark evaluation discussed in Section~\ref{subsec:liberopro_results}. While the main text presents averaged success rates across all tasks in each suite (Table~\ref{tab:libero_pro_summary}), the detailed tables below show individual task performance under each perturbation type. These results demonstrate the robustness characteristics of GLaD and UniVLA across specific manipulation scenarios, with particular emphasis on object appearance perturbations where GLaD shows significant advantages.

%


\begin{table*}[!t]
\centering
\caption{Detailed per-task results on LIBERO-Goal benchmark. Shows success rates (\%) for each task under five perturbation types: original (Ori), object appearance (Obj), position (Pos), semantic/language (Sem), and task composition (Task). Results averaged over 50 episodes per task.}
\label{tab:libero_goal_detailed}
\begin{tabular}{lcccccccccc}
\toprule
\multirow{2}{*}{Task} & \multicolumn{5}{c}{univla} & \multicolumn{5}{c}{GlaD} \\
\cmidrule(lr){2-6} \cmidrule(lr){7-11}
 & Ori & Obj & Pos & Sem & Task & Ori & Obj & Pos & Sem & Task \\
\midrule
Put(bowl, stove)                & 92 & 54 & 0 & 98 & 0 & \textbf{100} & \textbf{96} & \textbf{18} & \textbf{100} & 0 \\
Put(wine\_bottle, cabinet\_top) & 90 & 80 & 0 & 86 & 0 & \textbf{98} & \textbf{92} & 0 & \textbf{94} & 0 \\
Open(cabinet, drawer\_mid)      & \textbf{98} & \textbf{98} & 0 & 96 & 0 & 96 & 50 & 0 & \textbf{98} & \textbf{2} \\
TurnOn(stove)                   & 100 & 100 & 0 & 100 & 92 & 100 & 100 & 0 & 100 & \textbf{100} \\
Put(wine\_bottle, rack)         & 90 & \textbf{94} & 0 & 96 & 0 & \textbf{92} & 88 & 0 & 94 & 0 \\
Open(drawer\_top)  $\land$  Put(bowl, drawer\_top) & 88 & 20 & \textbf{36} & 90 & 0 & \textbf{96} & \textbf{62} & 26 & \textbf{98} & 0 \\
Push(plate, stove\_front)       & 98 & 100 & 0 & 100 & 0 & \textbf{100} & 100 & 0 & 100 & \textbf{2} \\
Put(bowl, plate)                & 98 & 24 & 0 & 100 & 0 & 98 & \textbf{84} & 0 & 100 & 0 \\
Put(bowl, cabinet\_top)         & 96 & 18 & 0 & 100 & 0 & \textbf{98} & \textbf{76} & 0 & 100 & 0 \\
Put(cream\_cheese, bowl)        & 100 & 30 & 0 & \textbf{100} & 0 & 100 & \textbf{64} & 0 & 98 & 0 \\
\midrule
Average & 95 & 62 & 4 & 97 & 9 & \textbf{98} & \textbf{81} & 4 & \textbf{98} & \textbf{10} \\
\bottomrule
\end{tabular}
\end{table*}


\begin{table*}[!t]
\centering
\caption{Detailed per-task results on LIBERO-Spatial benchmark. Shows success rates (\%) for spatial reasoning tasks. Results averaged over 50 episodes per task.}
\label{tab:libero_spatial_detailed}
\begin{tabular}{lcccccccccc}
\toprule
\multirow{2}{*}{Task} & \multicolumn{5}{c}{univla} & \multicolumn{5}{c}{GlaD} \\
\cmidrule(lr){2-6} \cmidrule(lr){7-11}
 & Ori & Obj & Pos & Sem & Task & Ori & Obj & Pos & Sem & Task \\
\midrule
Pick(on(cookie\_box), plate)            & 100 & 100 & 0 & 100 & --                      & 100 & 100 & 0 & 100 & -- \\
Pick(next\_to(ramekin), plate)          & 98 & 100 & 0 & 100 & --                       & \textbf{100} & 100 & 0 & 100 & -- \\
Pick(table\_center, plate)              & 98 & 100 & 0 & \textbf{100} & --              & 98 & 100 & \textbf{90} & 96 & -- \\
Pick(between(plate, ramekin), plate)    & 98 & 92 & 0 & 90 & --                         & \textbf{100} & \textbf{94} & 0 & 90 & -- \\
Pick(drawer\_top, plate)                & 92 & 96 & 0 & 94 & --                         & 92 & \textbf{100} & \textbf{30} & \textbf{100} & -- \\
Pick(next\_to(cookie\_box), plate)      & 100 & 100 & 0 & 100 & --                      & 100 & 100 & 0 & 100 & -- \\
Pick(next\_to(plate), plate)            & 100 & 92 & 0 & 88 & --                        & 100 & 92 & 0 & \textbf{90} & -- \\
Pick(on(ramekin), plate)                & 94 & 100 & 0 & 98 & --                        & \textbf{98} & 100 & 0 & \textbf{100} & -- \\
Pick(on(stove), plate)                  & \textbf{94} & 100 & 0 & 100 & --              & 92 & 100 & 0 & 100 & -- \\
Pick(on(cabinet), plate)                & \textbf{96} & \textbf{98} & 0 & 96 & --       & 88 & 96 & 0 & \textbf{98} & -- \\
\midrule
Average & 97 & 98 & 0 & 97 & -- & 97 & 98 & \textbf{12} & 97 & -- \\
\bottomrule
\end{tabular}
\end{table*}


\begin{table*}[!t]
\centering
\caption{Detailed per-task results on LIBERO-10 benchmark. Shows success rates (\%) for long-horizon multi-step manipulation tasks requiring complex sequences with multiple sub-goals. Results averaged over 50 episodes per task.}
\label{tab:libero_10_detailed}
\begin{tabular}{lcccccccccc}
\toprule
\multirow{2}{*}{Task} & \multicolumn{5}{c}{univla} & \multicolumn{5}{c}{GlaD} \\
\cmidrule(lr){2-6} \cmidrule(lr){7-11}
 & Ori & Obj & Pos & Sem & Task & Ori & Obj & Pos & Sem & Task \\
\midrule
Put({alphabet\_soup, tomato\_sauce}, basket)                                    & \textbf{98} & 0 & 0 & 94 & 0          & 96 & 2 & 0 & \textbf{98} & 0 \\
TurnOn(stove)  $\land$  Put(moka\_pot, stove)                                   & \textbf{100} & 92 & 0                 & \textbf{98} & 0 & 96 & \textbf{100} & 0 & 96 & 0 \\
Put(white\_mug, plate)  $\land$  Put(chocolate\_pudding, right\_of(plate))      & 88 & 42 & 0 & 88 & 0                  & \textbf{96} & \textbf{70} & 0 & \textbf{94} & 0 \\
Put(white\_mug, left\_plate)  $\land$  Put(yellow\_white\_mug, right\_plate)    & 82 & 74 & 0 & 74 & 0                  & \textbf86 & \textbf92 & 0 & \textbf88 & 0 \\
Put(black\_bowl, drawer\_bottom)  $\land$  Close(drawer\_bottom)                & 96 & \textbf88 & 0 & 92 & 0           & 96 & 86 & 0 & \textbf{96} & 0 \\
Put({cream\_cheese, butter}, basket)                                            & 96 & 96 & 0 & 98 & 0                  & 96 & \textbf{100} & 0 & \textbf{100} & 0 \\
Put({alphabet\_soup, cream\_cheese}, basket)                                    & \textbf{100} & 0 & 0 & 98 & 0         & 98 & 0 & 0 & \textbf{100} & 0 \\
Put({moka\_pot\_1, moka\_pot\_2}, stove)                                        & 78 & 58 & 0 & \textbf{72} & 88                 & 78 & \textbf{76} & 0 & 70 & 88 \\
Place(book, caddy\_back)                                                        & 96 & 0 & 12 & 98 & 0                  & \textbf{100} & 0 & \textbf{18} & \textbf{100} & 0 \\
Put(yellow\_white\_mug, microwave)  $\land$  Close(microwave)                   & \textbf{96} & \textbf{16} & 0 & \textbf{98} & 0                  & 94 & 14 & 0 & 88 & 0 \\
\midrule
Average & 93 & 47 & 1 & 91 & 9 & \textbf{94} & \textbf{54} & \textbf{2} & \textbf{93} & 9 \\
\bottomrule
\end{tabular}
\end{table*}


\begin{table*}[!t]
\centering
\caption{Detailed per-task results on LIBERO-Object benchmark. Shows success rates (\%) for object generalization tasks, testing placement of 10 different objects with varying visual properties. Results averaged over 50 episodes per task.}
\label{tab:libero_object_detailed}
\begin{tabular}{lcccccccccc}
\toprule
\multirow{2}{*}{Task} & \multicolumn{5}{c}{univla} & \multicolumn{5}{c}{GlaD} \\
\cmidrule(lr){2-6} \cmidrule(lr){7-11}
 & Ori & Obj & Pos & Sem & Task & Ori & Obj & Pos & Sem & Task \\
\midrule
Place(alphabet\_soup, basket)       & 92 & 8 & 0 & 98 & 0       & \textbf{100} & \textbf{26}& 0 & 98 & 0 \\
Place(bbq\_sauce, basket)           & \textbf{94} & 70 & 0 & \textbf{98} & 0      & 92 & \textbf{98} & \textbf{12} & 84 & 0 \\
Place(butter, basket)               & \textbf{98} & 96 & \textbf{4} & 94 & 0      & 96 & \textbf{100} & 0 & \textbf{100} & 0 \\
Place(chocolate\_pudding, basket)   & 100 & 98 & 0 & 100 & 0    & 100 & \textbf{100} & 0 & 100 & 0 \\
Place(cream\_cheese, basket)        & 94 & 100 & \textbf{36} & 98 & 0    & \textbf{98} & 100 & 22 & \textbf{100} & 0 \\
Place(ketchup, basket)              & 98 & \textbf{74} & 0 & 98 & 0      & 98 & 38 & 0 & 98 & 0 \\
Place(milk, basket)                 & \textbf{98} & 78 & 0 & \textbf{100} & 0     & 96 & \textbf{98} & 0 & 94 & 0 \\
Place(orange\_juice, basket)        & 88 & 98 & 0 & 96 & 0      & \textbf{100} & \textbf{100} & 0 & \textbf{100} & 0 \\
Place(salad\_dressing, basket)      & 96 & 98 & 0 & 96 & 0      & \textbf{100} & \textbf{100} & 0 & \textbf{98} & 0 \\
Place(tomato\_sauce, basket)        & \textbf{100} & 100 & 0 & 96 & 0    & 92 & 100 & 0 & \textbf{100} & 0 \\
\midrule
Average & 96 & 82 & \textbf{4} & 97 & 0 & \textbf{97} & \textbf{86} & 3 & 97 & 0 \\
\bottomrule
\end{tabular}
\end{table*}

\end{document}